\documentclass[11pt,a4paper]{article}
\usepackage[hyperref]{emnlp-ijcnlp-2019}

\usepackage{times}
\usepackage{latexsym}
\usepackage{microtype}
\usepackage{natbib}
\usepackage{booktabs}
\usepackage{appendix}
\usepackage{tikz}
\usepackage{tikz-dependency}
\usepackage{qtree}
\usepackage{amsmath}
\usepackage{amsfonts}
\usepackage{graphicx}
\usepackage{amsthm}
\usepackage{mathtools}
\usepackage{multirow}
\usepackage{url}
\usepackage{color}
\usepackage[smallerops]{newtxmath}
\usepackage{subfig}

\usepackage[disable]{todonotes}
\makeatletter
\newcommand*\iftodonotes{\if@todonotes@disabled\expandafter\@secondoftwo\else\expandafter\@firstoftwo\fi}  
\makeatother

\makeatletter
\newcommand\footnoteref[1]{\protected@xdef\@thefnmark{\ref{#1}}\@footnotemark}
\makeatother
\newcommand{\noindentaftertodo}{\iftodonotes{\noindent}{}\ignorespaces}
\newcommand{\note}[4][]{\todo[author=#2,color=#3,size=\scriptsize,fancyline,caption={},#1]{#4}} 
\newcommand{\jason}[2][]{\note[#1]{jason}{green!40}{#2}}
\newcommand{\lisa}[2][]{\note[#1]{lisa}{orange!40}{#2}}

\newcommand{\Jason}[2][]{\jason[inline,#1]{#2}\noindentaftertodo}

\newlength{\extramargin}
\usepackage{calc}
\iftodonotes{\usepackage[paperwidth=\paperwidth+\extramargin*2,marginparwidth=\marginparwidth+\extramargin,width=\textwidth,height=\textheight]{geometry}}{}

\definecolor{oracle}{rgb}{0.5, 0.5, 0.5}
\newcommand{\grey}[1]{{\color{oracle}#1}}

\usepackage[noabbrev,capitalize]{cleveref} 

\crefname{equation}{equation}{equations}   
\crefname{footnote}{footnote}{footnotes}   
\crefname{line}{line}{lines}               
\crefname{section}{\S}{\S\S}
\Crefname{section}{\S}{\S\S}    
\crefformat{section}{#2\S#1#3}  
\Crefformat{section}{#2\S#1#3}
\crefrangeformat{section}{\S\S#3#1#4--#5#2#6}
\Crefrangeformat{section}{\S\S#3#1#4--#5#2#6}

\usepackage{appendix}

\newcommand{\defeq}{\mathrel{\stackrel{\textnormal{\tiny def}}{=}}} 
\newcommand{\E}[2][]{\mathop{\mathbb{E}}_{{#1}}\,[#2]}   
\newcommand{\Ebare}[1][]{\mathop{\mathbb{E}}_{{#1}}}
\newcommand{\Ee}[2][]{\smashoperator[r]{\mathop{\mathbb{E}}_{{#1}}}\,[#2]}   
\DeclareMathOperator{\MI}{I}
\DeclareMathOperator{\KL}{KL}

\DeclareMathOperator{\tr}{tr}
\DeclareMathOperator*{\argmax}{argmax}

\DeclareMathOperator{\Entr}{H}

\newcommand{\ptheta}{p_\theta}
\newcommand{\qphi}{q_\phi}
\newcommand{\rpsi}{r_\psi}
\newcommand{\sxi}{s_\xi}
\newcommand{\xtype}{\hat{x_i}}
\newcommand{\xType}{\hat{X_i}}
\DeclareCaptionLabelFormat{andtable}{#1~#2  \&  \tablename~\thetable}

\aclfinalcopy 
\def\aclpaperid{1357} 

\title{Specializing Word Embeddings (for Parsing) by Information Bottleneck}

\author{Xiang Lisa Li \\
Department of Computer Science\\ 
Johns Hopkins University \\
  \texttt{xli150@jhu.edu} \\\And
  Jason Eisner \\
  Department of Computer Science\\ 
Johns Hopkins University \\
  \texttt{jason@cs.jhu.edu} \\}

\date{}

\begin{document}
\maketitle
\Jason{Don't forget to deanonymize and switch to the final-version style!  Be sure to follow all camera-ready requirements.  We also need to add acknowledgments, including to PURA (?) and NSF.}
\begin{abstract}
  Pre-trained word embeddings like ELMo and BERT contain rich syntactic 
  and semantic information, resulting in state-of-the-art performance on 
  various tasks.  We propose a very fast variational information bottleneck (VIB) method to nonlinearly compress these embeddings, keeping only the information that helps a discriminative parser. We compress each word embedding to either a discrete tag or a continuous vector. 
  In the \emph{discrete} version, our automatically compressed tags form an alternative tag set: we show experimentally that our tags capture most of the information in traditional POS tag annotations, but our tag sequences can be parsed more accurately at the same level of tag granularity.
   In the \emph{continuous} version, we show experimentally that moderately compressing the word embeddings by our method yields a more accurate parser in 8 of 9 languages, unlike simple dimensionality reduction.



\end{abstract}

\section{Introduction}\label{sec:intro}
Word embedding systems like BERT and ELMo use spelling and context to obtain contextual embeddings of word tokens.  These systems are trained on large corpora in a task-independent way.  The resulting embeddings have proved to then be useful for both syntactic and semantic tasks, with different layers of ELMo or BERT being somewhat specialized to different kinds of tasks \cite{BiLM-dissect,BERT-syntax}.  State-of-the-art performance on many NLP tasks can be obtained by fine-tuning, i.e., back-propagating task loss all the way back into the embedding function \cite{ELMo-orig,BERT-orig}.
\begin{figure}
  \includegraphics[page=1,width=0.48\textwidth]{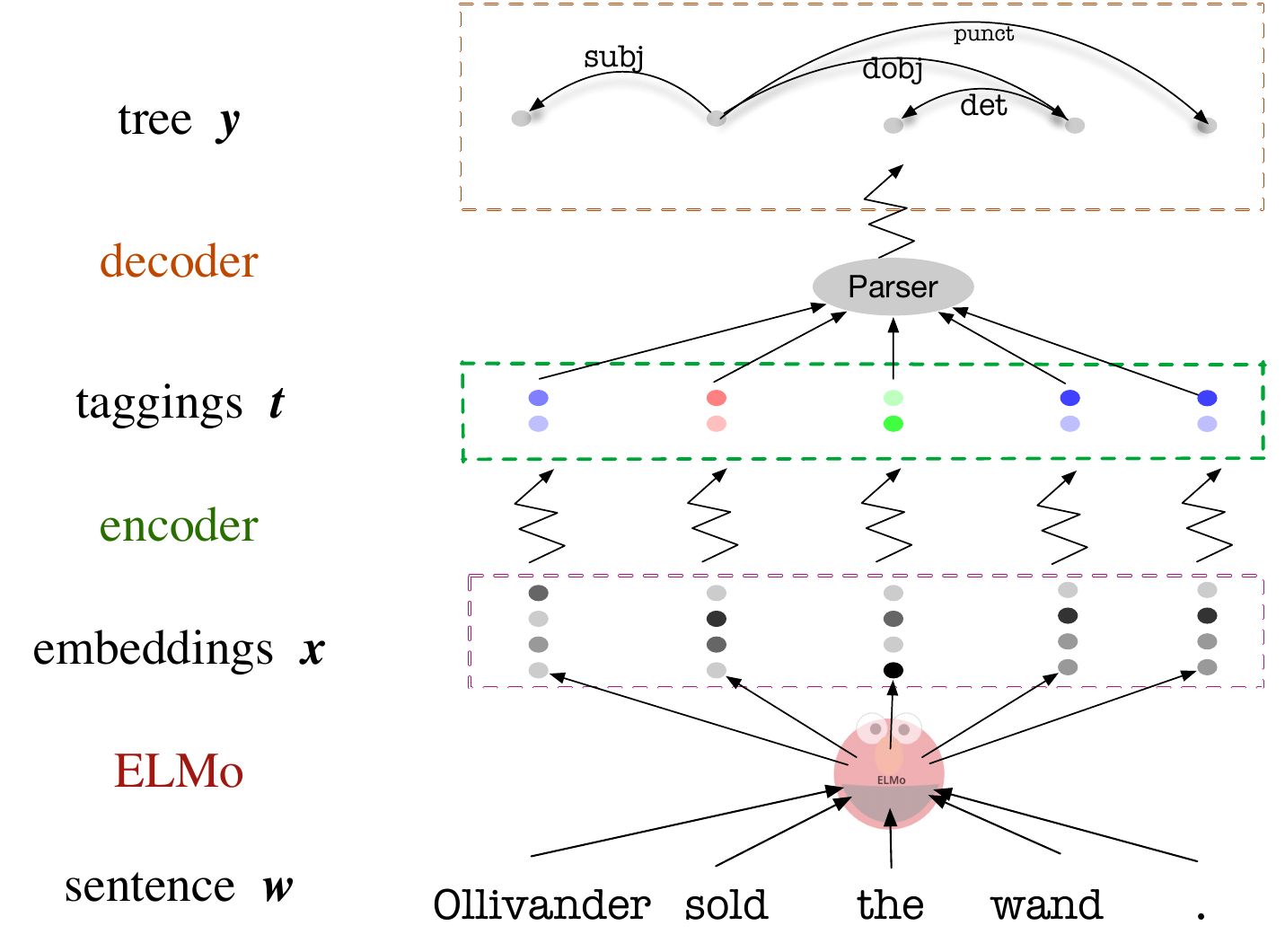}
  \caption{ \label{fig:graph_model} Our instantiation of the information bottleneck, with bottleneck variable $T$.  A jagged arrow indicates a stochastic mapping, i.e. the jagged arrow points from the parameters of a distribution to a sample drawn from that distribution.}
\end{figure}

In this paper, we explore what task-specific information appears in the embeddings \emph{before} fine-tuning takes place.  We focus on the task of dependency parsing, but our method can be easily extended to other syntactic or semantic tasks.
Our method compresses the embeddings by extracting just their syntactic properties---specifically, the information needed to reconstruct parse trees (because that is our task).  Our nonlinear, stochastic compression function is explicitly trained by variational information bottleneck (VIB) to forget task-irrelevant information.  This is reminiscent of canonical correspondence analysis \cite{anderson2003introduction}, a method for reducing the dimensionality of an input vector so that it remains predictive of an output vector, although we are predicting an output tree instead. \Jason{Should we in fact try some kind of deep CCA (Arora) as another baseline?  In other words (I think), don't use randomness or information-theoretic measures controlled by $\beta$: just choose a deterministic function that projects down to some smaller tagset or lower-dimensional vector.}  However, VIB goes beyond mere dimensionality reduction to a fixed lower dimensionality, since it also avoids unnecessary \emph{use} of the dimensions that are available in the compressed representation, blurring unneeded capacity via randomness.  The effective number of dimensions may therefore vary from token to token.  For example, a parser may be content to know about an adjective token only that it is adjectival, whereas to find the dependents of a verb token, it may need to know the verb's number and transitivity, and to attach a preposition token, it may need to know the identity of the preposition.

We try compressing to both discrete and continuous task-specific representations.  Discrete representations yield an interpretable clustering of words.  
We also extend information bottleneck to allow us to control the contextual specificity of the token embeddings, making them more like type embeddings.  

This specialization method is complementary to the previous fine-tuning approach. Fine-tuning introduces \emph{new} information into word embeddings by backpropagating the loss, whereas the VIB method learns to exploit the \emph{existing} information found by the ELMo or BERT language model.  VIB also has less capacity and less danger of overfitting, since it fits fewer parameters than fine-tuning (which in the case of BERT has the freedom to adjust the embeddings of all words and word pieces, even those that are rare in the supervised fine-tuning data).  VIB is also very fast to train on a single GPU.

We discover that our syntactically specialized embeddings are predictive of the gold POS tags in the setting of few-shot-learning, validating the intuition that a POS tag summarizes a word token's \emph{syntactic} properties.  However, our representations are tuned explicitly for discriminative parsing, so they prove to be even more useful for this task than POS tags, even at the same level of granularity.  They are also more useful than the uncompressed ELMo representations, when it comes to generalizing to test data.  (The first comparison uses discrete tags, and the second uses continuous tags.)

\section{Background: Information Bottleneck}
The information bottleneck (IB) method originated in information theory and has been adopted by the machine learning community as a training objective \cite{tishby2000information} and a theoretical framework for analyzing deep neural networks \cite{Tishby2015DeepLA}. 

Let $X$ represent an ``input'' random variable such as a sentence, and $Y$ represent a correlated ``output'' random variable such as a parse.  Suppose we know the joint distribution $p(X,Y)$. (In practice, we will use the empirical distribution over a sample of $(x,y)$ pairs.)  Our goal is to learn a stochastic map $\ptheta(t\mid x)$ from $X$ to some compressed representation $T$, which in our setting will be something like a tag sequence.  IB seeks to minimize 

\noindent
\vspace{-\baselineskip}
\begin{equation}
\label{eq:Obj}
\mathcal{L}_{IB} = -\MI(Y;T) + \beta \cdot \MI(X;T)
\end{equation}
where $\MI(\cdot; \cdot)$ is the mutual
information.\footnote{In our IB notation, larger $\beta$ means more compression. Note that there is another version of IB that puts $\beta$ as the coefficient in front of $\MI(Y;T)$: $\mathcal{L}_{IB} = -\beta\cdot\MI(Y;T) + \MI(X;T)$ The two versions are equivalent.}  A low loss means that $T$ does not retain very much information about $X$ (the second term), while still retaining enough information to predict $Y$.\footnote{ \label{IB_indep} Since $T$ is a stochastic function of $X$ with no access to $Y$, it obviously cannot convey more information about $Y$ than the uncompressed input $X$ does.  As a result, $Y$ is independent of $T$ given $X$, as in the graphical model $T \rightarrow X \rightarrow Y$.} 
The balance between the two MI terms is controlled by a Lagrange multiplier $\beta$. By increasing $\beta$, we increase the pressure to keep $\MI(X;T)$ small, which ``narrows the bottleneck'' by favoring compression over predictive accuracy $\MI(Y;T)$.  Regarding $\beta$ as a Lagrange multiplier, we see that the goal of IB is to maximize the predictive power of $T$ subject to some constraint on the amount of information about $X$ that $T$ carries.  If the map from $X$ to $T$ were deterministic, then it could lose information only by being non-injective: the traditional example is dimensionality reduction, as in the encoder of an encoder-decoder neural net.  But IB works even if $T$ can take values throughout a high-dimensional space, because the randomness in $\ptheta(t\mid x)$ means that $T$ is noisy in a way that wipes out information about $X$.  Using a high-dimensional space is desirable because it permits the amount of effective dimensionality reduction to vary, with $T$ perhaps retaining much more information about some $x$ values than others, as long as the \emph{average} retained information $\MI(X;T)$ is small.


\section{Formal Model}

\label{sec:formal_model}

In this paper, we extend the original IB objective \eqref{eq:Obj} and 
add terms $\MI(T_i; X|\xType)$ to control the context-sensitivity of the
extracted tags.  Here $T_i$ is the tag associated with the $i$th word, $X_i$ is the ELMo token embedding of the $i$th word, and $\xType$ is the same word's ELMo type embedding (before context is incorporated).
 \begin{equation}  
 \label{eq:Obj3}
\mathcal{L}_{IB} = -\MI(Y;T) + \beta \MI(X;T) + \gamma \sum_{i=1}^n\MI(T_i;X|\xType)
\end{equation}
In this section, we will explain the motivation for the additional term and how to efficiently estimate variational bounds on all terms (lower bound for $\MI(Y;T)$ and upper bound for the rest).
\footnote{\label{fn:diffent}Traditional Shannon entropy $\Entr(\cdot)$ is defined on discrete variables. In the case of continuous variables, we interpret $\Entr$ to instead denote differential entropy (which would be $-\infty$ for discrete variables).  Scaling a continuous random variable affects its differential entropy---but not its mutual information with another random variable, which is what we use here.
}

We instantiate the variational IB (VIB) estimation method \cite{DVIB} on our dependency parsing task, as illustrated in \cref{fig:graph_model}. We compress a sentence's word embeddings $X_i$ into continuous vector-valued tags or discrete tags $T_i$ (``encoding'') such that the tag sequence $T$ retains maximum ability to predict the dependency parse $Y$ (``decoding'').  Our chosen architecture compresses each $X_i$ independently using the same stochastic, information-losing transformation.  


The IB method introduces the new random variable $T$, the tag sequence that compresses $X$, by defining the conditional distribution $\ptheta(t\mid x)$. In our setting, $\ptheta$ is a stochastic tagger, for which we will adopt a parametric form (\cref{sec:encoder} below).   Its parameters $\theta$ are chosen to minimize the IB objective \eqref{eq:Obj3}. By IB's independence assumption,\footnoteref{IB_indep} the joint probability can be factored as $\ptheta(x,y,t) = p(x) \cdot p(y \mid x) \cdot \ptheta(t\mid x)$.


\subsection{$\MI(X;T)$ --- the Token Encoder $\ptheta(t\mid x)$}
\label{sec:encoder}
Under this distribution, $\MI(X;T) \!\defeq\! \E[x,t]{\log \frac{\ptheta(t|x)}{\ptheta(t)}} 
= \E[x]{\E[t \sim \ptheta(t | x)]{\log \frac{\ptheta(t|x)}{\ptheta(t)}}}$. Making this term small yields a representation $T$ that, on average, retains little information about $X$.
The outer expectation is over the true distribution of sentences $x$; we use an empirical estimate, averaging over the unparsed sentences in a dependency treebank. To estimate the inner expectation, we could sample,
drawing taggings $t$ from $\ptheta(t \mid x)$.

We must also compute the quantities within the inner brackets. The $\ptheta(t\mid x)$ term is defined by our parametric form.  The troublesome term is $\ptheta(t) =\E[x']{\ptheta(t\mid x')}$, since even estimating it from a treebank requires an inner loop over treebank sentences $x'$.  To avoid this, variational IB replaces $\ptheta(t)$ with some variational distribution $\rpsi(t)$.  This can only increase our objective function, since the difference between the variational and original versions of this term is a KL divergence and hence non-negative:
%
%
\begin{equation}
\begin{split}
& \overbrace{  \Ee[x]{\Ee[\substack{ t \sim \ptheta(t\mid x)}]{\log \frac{\ptheta(t|x)}{\rpsi(t)}}} }^\text{upper bound} - 
\overbrace{  \Ee[x] {\Ee[\substack{ t \sim \ptheta(t\mid x)}] {\log \frac{\ptheta(t\mid x)}{\ptheta(t)}}} }^{\MI(X;T)}\\ \nonumber
& = \Ee[x] {\KL (\ptheta(t) \mid \mid \rpsi(t))}  \geq 0
\end{split}
\end{equation}
Thus, the variational version (the first term above) is indeed an upper bound for $\MI(X;T)$ (the second term above).  We will minimize this upper bound by 
adjusting not only $\theta$ but also $\psi$, thus making
the bound as tight as possible given $\theta$.  Also we will no longer need to sample $t$ for the inner expectation of the upper bound, $\E[\substack{ t \sim \ptheta(t\mid x)}] {\log \frac{\ptheta(t\mid x)}{\rpsi(t)}}$, because this expectation equals $\KL[ \ptheta(t\mid x)\mid\mid \rpsi(t) ]$, and we will define the parametric $\ptheta$ and $\rpsi$ so that this $\KL$ divergence can be computed exactly: see \cref{sec:train}.

\subsection{Two Token Encoder Architectures}\label{sec:two-encoders}
We choose to define $\ptheta(t\mid x) = \prod_{i=1}^n \ptheta(t_i\mid x_i)$.  That is, our stochastic encoder will compress each word $x_i$ individually (although $x_i$ is itself a representation that depends on context): see \cref{fig:graph_model}.
We make this choice not for computational reasons---our method would remain tractable even without this---but because our goal in this paper is to find the syntactic information in each individual ELMo token embedding (a goal we will further pursue in \cref{sec:context} below).

To obtain continuous tags, define $\ptheta(t_i \mid x_i)$ such that $t_i \in \mathbb{R}^d$ is Gaussian-distributed with mean vector and diagonal covariance matrix computed from the ELMo word vector $x_i$ via a feedforward neural network with $2d$ outputs and no transfer function at the output layer.  To ensure positive semidefiniteness of the diagonal covariance matrix, we squared the latter $d$ outputs to obtain the diagonal entries.\footnote{Our restriction to diagonal covariance matrices follows \citet{DVIB}.  In pilot experiments that dropped this restriction, we found learning to be numerically unstable, although that generalization is reasonable in principle.}

Alternatively, to obtain discrete tags, define $\ptheta(t_i \mid x_i)$ such that $t_i \in \{1,\ldots,k\}$ follows a softmax distribution, where the $k$ softmax parameters are similarly computed by a feedforward network with $k$ outputs and no transfer function at the output layer.

We similarly define $\rpsi(t) = \prod_{i=1}^n \rpsi(t_i)$, where $\psi$ directly specifies the $2d$ or $k$ values corresponding to the output layer above (since there is no input $x_i$ to condition on).


\subsection{$\MI(T_i; X|\xType)$ --- the Type Encoder $\sxi(t_i|\xtype)$}
\label{sec:context}
While the IB objective \eqref{eq:Obj} asks each tag $t_i$ to be informative about the parse $Y$, we were concerned that it might not be interpretable as a tag of word $i$ specifically.  Given ELMo or any other black-box conversion of a length-$n$ sentence to a sequence of contextual vectors $x_1, \ldots, x_n$, it is possible that $x_i$ contains not only information about word $i$ but also information describing word $i+1$, say, or the syntactic constructions in the vicinity of word $i$.  Thus, while $\ptheta(t_i \mid x_i)$ might extract some information from $x_i$ that is very useful for parsing, there is no guarantee that this information came from word $i$ and not its neighbors.  Although we \emph{do} want tag $t_i$ to consider context---e.g., to distinguish between noun and verb uses of word $i$---we want ``most'' of $t_i$'s information to come from word $i$ itself.  Specifically, it should come from ELMo's level-0 embedding of word $i$, denoted by $\xtype$---a word \emph{type} embedding that does \emph{not} depend on context.


To penalize $T_i$ for capturing ``too much'' contextual information, our modified objective \eqref{eq:Obj3} adds a penalty term $\gamma \cdot \MI(T_i; X|\xType)$, which measures the amount of information about $T_i$ given by the sentence $X$ as a whole, beyond what is given by $\xType$:
$\MI(T_i;X\!\mid\!\xType)\!\defeq\!\E[x]{\E[t_i \sim \ptheta(t_i\mid x)]{\log \frac{\ptheta(t_i\mid x)}{\ptheta(t_i\mid \xtype)}}}$.
Setting $\gamma > 0$ will reduce this contextual information.  

In practice, we found that $\MI(T_i; X\mid\xType)$ was small even when $\gamma=0$, on the order of 3.5 nats whereas $\MI(T_i;X)$ was 50 nats.  In other words, the tags extracted by the classical method were already fairly local, so increasing $\gamma$ above 0 had little qualitative effect.  Still, $\gamma$ might be important when applying our method to ELMo's competitors such as BERT.

We can derive an upper bound on $\MI(T_i;X|\xType)$ by approximating the conditional distribution $\ptheta(t_i \mid \xtype)$ with a variational distribution $\sxi(t_i \mid \xtype)$, similar to \cref{sec:encoder}.  
\vspace{-10pt}
\begin{equation}
\begin{split}
& \overbrace{\E[x]{\Ee[\substack{ t_i \sim  \ptheta(t_i\mid x)}]{\log \frac{\ptheta(t_i|x)}{\sxi(t_i |\xtype)}}}}^\text{upper bound} - 
\overbrace{\Ee[x]{ \Ee [\substack{ t_i \sim   \ptheta(t_i\mid x)}] {\log \frac{\ptheta(t_i| x)}{\ptheta(t_i|\xtype)}}}}^{\MI(T_i;X|\xType)}\\ \nonumber
& = \E[x] {\KL (\ptheta(t_i \mid \xtype) \mid \mid \sxi(t_i \mid \xtype))} \geq 0
\end{split}
\end{equation}
We replace it in \eqref{eq:Obj3} with this upper bound, which is equal to 
$\E[x]{\sum_{i=1}^n \KL[\ptheta(t_i|x) \mid\mid \sxi(t_i \mid \xtype)}]$.

The formal presentation above does not assume the specific factored model that we adopted in \cref{sec:two-encoders}.  When we adopt that model, $\ptheta(t_i \mid x)$ above reduces to $\ptheta(t_i \mid x_i)$---but our method in this section still has an effect, because $x_i$ still reflects the context of the full sentence whereas $\xtype$ does not.  

\paragraph{Type Encoder Architectures}
Notice that $\sxi(t_i \mid \xtype)$ may be regarded as a type encoder, with parameters $\xi$ that are distinct from the parameters $\theta$ of our token encoder $\ptheta(t_i \mid x_i)$. 
Given a choice of neural architecture for $\ptheta(t_i \mid x_i)$ (see \cref{sec:two-encoders}), we always use the same architecture for $\sxi (t_i \mid \xtype)$, except that $\ptheta$ takes a token vector as input whereas $\sxi$ takes a context-independent type vector.  $\sxi$ is not used at test time, but only as part of our training objective.


\subsection{$\MI(Y;T)$ --- the Decoder $\qphi(y\mid t)$}
\label{sec:decoder}
Finally, $\MI(Y;T)\!\defeq\! \E[y,t \sim \ptheta]{\log \frac{\ptheta(y|t)}{p(y)}}$.  The $p(y)$ can be omitted during optimization as it does not depend on $\theta$. Thus, making $\MI(Y;T)$ large tries to obtain a high log-probability $\ptheta(y\mid t)$ for the true parse $y$ when reconstructing it from $t$ alone.  

But how do we compute $\ptheta(y\mid t)$?  This quantity effectively marginalizes over possible sentences $x$ that {\em could have} explained $t$.
Recall that $\ptheta$ is a joint distribution over $x,y,t$: see just above \cref{sec:encoder}.  So $\ptheta(y\mid t) \defeq \frac{\sum_{x} \ptheta(x,y,t)}{\sum_{x,y'} \ptheta(x,y',t)}$.
To estimate these sums accurately, we would have to identify the sentences $x$ that are most consistent with the tagging $t$ (that is, $p(x)\cdot \ptheta(t|x)$ is large): these contribute the largest summands, but might not appear in any corpus. 

To avoid this, we replace $\ptheta(y\mid t)$ with a variational approximation $\qphi(y\mid t)$ in our formula for $\MI(Y;T)$.  Here $\qphi(\cdot \mid \cdot)$ is a tractable conditional distribution, and may be regarded as a stochastic parser that runs on a compressed tag sequence $t$ instead of a word embedding sequence $x$.  This modified version of $\MI(Y;T)$ forms a lower bound on $\MI(Y;T)$, for any value of the variational parameters $\phi$, since the difference between them is a KL divergence and hence positive:
\vspace{-10pt}
\begin{equation}
\begin{split}
&\overbrace{\Ee[y,t \sim \ptheta]{ \log \tfrac{\ptheta(y \mid t) } {p(y)}} } ^{\MI(Y;T)} -  \overbrace{\Ee[y,t \sim \ptheta]{ \log \tfrac{\qphi(y \mid t)}{p(y)}}}^\text{lower bound} \\
&= \E[t \sim \ptheta]{\KL(\ptheta(y \mid t) \mid \mid \qphi(y \mid t))} \geq 0 \nonumber
\end{split}
\end{equation}
\vspace{-2pt}
We will maximize this lower bound of $\MI(Y;T)$ with respect to both $\theta$ and $\phi$. For any given $\theta$, the optimal $\phi$ minimizes the expected KL divergence, meaning that $\qphi$ approximates $\ptheta$ well.

More precisely, we again drop $p(y)$ as constant and then maximize a \emph{sampling-based estimate} of $\E[y,t \sim \ptheta]{ \log \qphi(y|t)}$.  To sample $y,t$ from the joint $\ptheta(x,y,t)$ we must first sample $x$, so we rewrite as $\E[x,y] { \E[t \sim \ptheta(t \mid x)] { \log \qphi(y|t)}}$.  
%
%
The outer expectation $\Ebare[x,y]$ is estimated as usual over a training treebank.  The expectation $\Ebare[t \sim \ptheta(t\mid x)]$ recognizes that $t$ is stochastic, and again we estimate it by sampling.  In short, when $t$ is a stochastic compression of a treebank sentence $x$, we would like our variational parser \emph{on average} to assign high log-probability $\qphi(y\mid t)$ to its treebank parse $y$.

\paragraph{Decoder Architecture}

We use the deep biaffine dependency parser \cite{DBLP:journals/corr/DozatM16} as our variational distribution $\qphi(y\mid t)$, which functions as the decoder. This parser uses a Bi-LSTM to extract features from compressed tags or vectors and assign scores to each tree edge, setting $\qphi(y \mid t)$ proportional to the $\exp$ of the total score of all edges in $y$.  During IB training, the code\footnote{We use the implementation from AllenNLP library \cite{AllenNLP}.} computes only an approximation to $\qphi(y|t)$ for the gold tree $y$ (although in principle, it could have computed the exact normalizing constant in polytime with Tutte's matrix-tree theorem \cite{smith2007probabilistic,koo2007structured,mcdonald2007complexity}).  When we test the parser, the code does exactly find $\argmax_y \qphi(y\mid t)$ via the directed spanning tree algorithm of \citet{edmond}.


\section{Training and Inference}
\label{sec:train}

\lisa{META: it would be good to use the extra space to make section 4 more self-contained by adding more explanation.}
With the approximations in \cref{sec:formal_model}, our final minimization objective is this upper bound on \eqref{eq:Obj3}:
  \begin{align}
   \label{eq:combine}
 \Ebare[x,y] \Big[ & \Ee[t \sim  \ptheta(t\mid x)] {{\scriptstyle-} \log \qphi(y|t)} +  \beta \KL (\ptheta(t|x) || \rpsi(t)) \nonumber \\ 
  & + \gamma \sum_{i=1}^{n} \KL(\ptheta(t_i \mid x) \mid \mid \sxi(t_i\mid \xtype)) \;  \Big]
  \end{align}
We apply stochastic gradient descent to optimize this objective. To get a stochastic estimate of the objective, 
we first sample some $(x,y)$ from the treebank.  We then have many expectations over $t \sim \ptheta(t \mid x)$, including the KL terms.  We could estimate these by sampling $t$ from the token encoder $\ptheta(t \mid x)$ and then evaluating all $\qphi, \ptheta, \rpsi$, and $\sxi$ probabilities.  However, in fact we use the sampled $t$ only to estimate the first expectation (by computing the decoder probability $\qphi(y \mid t)$ of the gold tree $y$); we can compute the KL terms exactly by exploiting the structure of our distributions.  The structure of $\ptheta$ and $\rpsi$ means that the first KL term decomposes into
$\sum_{i=1}^n \KL (\ptheta(t_i|x_i) || \rpsi(t_i))$.  All KL terms are now between either two Gaussian distributions over a continuous tagset\footnote{$\KL(\mathcal{N}_0 \mid \mid \mathcal{N}_1) = \frac{1}{2} (\tr(\Sigma_1^{-1} \Sigma_0) + (\mu_1 - \mu_0)^T \Sigma_1^{-1} (\mu_1 - \mu_0) - d + \log(\frac{\det(\Sigma_1)}{\det(\Sigma_0)})$} 
or two categorical distributions over a small discrete tagset.\footnote{ $\KL (\ptheta(t_i|x_i) || \rpsi(t_i)) = \sum_{t_i=1}^k \ptheta(t_i\mid x_i) \log \frac{\ptheta(t_i \mid x_i)}{\rpsi(t_i)} $} 

To compute the stochastic gradient, we run back-propagation on this computation. We must apply the reparametrization trick to backpropagate through the step that sampled $t$.  This finds the gradient of parameters that derive $t$ from a random variate $z$, while holding $z$ itself fixed.  For continuous $t$, we use the reparametrization trick for multivariate Gaussians \cite{rezende2014stochastic}. For discrete $t$, we use the Gumbel-softmax variant \cite{gumbel, MaddisonMT16}.


To evaluate our trained model's ability to parse a sentence $x$ from compressed tags, we obtain a parse as $\argmax_y \qphi(y\mid t)$, where $t \sim \ptheta(\cdot \mid x)$ is a single sample.  A better parser would instead estimate $\argmax_y \E[t]{\qphi(y\mid t)}$ where $\Ebare[t]$ averages over many samples $t$, but this is computationally hard.\looseness=-1

\section{Experimental Setup}

\paragraph{Data}
\newcommand{\tbf}[1]{\textsf{#1}}
Throughout \crefrange{sec:intrinsic}{sec:extrinsic}, we will examine our compressed tags on a subset of Universal Dependencies \cite{UNIVDEP-2.3}, or UD, a collection of dependency treebanks across 76 languages using the same POS tags and dependency labels.  We experiment on Arabic, Hindi, English, French, Spanish, Portuguese, Russian, Italian, and Chinese (\cref{tb:data})---languages with different syntactic properties like word order.  We use only the sentences with length $\leq 30$.  For each sentence, $x$ is obtained by running the standard pre-trained ELMo on the UD token sequence (although UD's tokenization may not perfectly match that of ELMo's training data), and $y$ is the labeled UD dependency parse \emph{without} any part-of-speech (POS) tags.  Thus, our tags $t$ are tuned to predict only the dependency relations in UD, and not the gold POS tags $a$ also in UD.\looseness=-1

\begin{table}
\setlength\tabcolsep{2pt}
\centering
\resizebox{0.4\textwidth}{!}{
\begin{tabular}{l l r c c}
\toprule
    Language             &Treebank    & \#Tokens &  $\Entr(A\mid \hat{X})$ & $\Entr(A)$ \\\midrule
    Arabic               &\small\tbf{PADT}  &    282k    &    0.059   & 2.059      \\
    Chinese              &\small\tbf{GSD}  &    123k    &    0.162  & 2.201 \\
    English              &\small\tbf{EWT}  &    254k    &    0.216  & 2.494  \\
    French               &\small\tbf{GSD}  &     400k  & 0.106 &2.335\\
    Hindi                &\small\tbf{HDTB}  &    351k    &    0.146   & 2.261   \\
    Portuguese           & \small\tbf{Bosque} &     319k & 0.179 & 2.305\\
    Russian               & \small\tbf{GSD} &     98k  & 0.049 & 2.132\\
    Spanish              &\small\tbf{AnCora}  &    549k    &  0.108  & 2.347   \\
    Italian & \small\tbf{ISDT}& 298K& 0.120 &2.304 \\
    \bottomrule
\end{tabular}}
\vspace{-2pt}
\caption{\label{tb:data}Statistics of the datasets used in this paper. ``Treebank'' is the treebank identifier in UD, ``\#Token'' is the number of tokens in the treebank, ``$\Entr(A)$'' is the entropy of a gold POS tag (in nats), and ``$\Entr(A \mid \hat{X})$'' is the conditional entropy of a gold POS tag conditioned on a word type (in nats).}
\end{table}

\paragraph{Pretrained Word Embeddings}
For English, we used the pre-trained English ELMo model from the AllenNLP library \cite{AllenNLP}. %
For the other 8 languages, we used the pre-trained models from \newcite{ELMo-Many-Lang}.
%
Recall that ELMo has two layers of bidirectional LSTM (layer 1 and 2) built upon a context-independent character CNN (layer 0). 
We use either layer 1 or 2 as the input ($x_i$) to our token encoder $\ptheta$.  Layer 0 is the input ($\xtype$) to our type encoder $\sxi$. 
Each encoder network (\crefrange{sec:two-encoders}{sec:context}) has a single hidden layer with a $\tanh$ transfer function, which has $2d$ hidden units (typically 128 or 512) for continuous encodings and 512 hidden units for discrete encodings.  

\paragraph{Optimization}
We optimize with Adam \cite{kingma2014adam}, a variant of stochastic gradient descent. We alternate between improving the model $\ptheta(t|x)$ on even epochs and the variational distributions $\qphi(y|t)$, $\rpsi(t)$, $\sxi(t_i\mid \xtype)$ on odd epochs. 

We train for 50 epochs with minibatches of size $20$ and L$_2$ regularization.  The learning rate and the regularization coefficients are tuned on dev data for each language separately. For each training sentence, we average over $5$ i.i.d.\@ samples of $T$ to reduce the variance of the stochastic gradient. \jason{is this theoretically motivated?} \lisa{I forgot why we ask this question, is this about why averaging over 5 iid samples can reduce variance? as shown in the paper, Miller at el. Reducing Reparameterization Gradient Variance}  The initial parameters $\theta, \phi, \psi, \xi$ are all drawn from $\mathcal{N}(0,I)$. We experiment with different dimensionalities $d \in \{5, 32,256, 512\}$ for the continuous tags, and different cardinalities $k \in \{32, 64, 128\}$  for the discrete tag set. We also tried different values $\beta, \gamma \in \{ 10^{-6}, 10^{-5}, \cdots, 10^1 \}$ of the compression tradeoff parameter. 
We use temperature annealing when sampling from the Gumbel-softmax distribution (\cref{sec:train}).  At training epoch $i$, we use temperature $\tau_i$, where $\tau_1=5$ and $\tau_{i+1}=\max ( 0.5, e^{-\gamma}\tau_i )$. We set the annealing rate $\gamma=0.1$. During testing, we use $\tau = 0$, which gives exact softmax sampling.

\section{Scientific Evaluation}
\label{sec:intrinsic}
In this section, we study what information about words is retained by our automatically constructed tagging schemes.
First, we show the relationship between $I(Y;T)$ and $I(X;T)$ on English as we reduce $\beta$ to capture more information in our tags.\footnote{We always set $\gamma = \beta$ to simplify the experimental design.} 

Second, across 9 languages, we study how our automatic tags correlate with gold part-of-speech tags (and in English, with other syntactic properties), while suppressing information about semantic properties. 
We also show how decreasing $\beta$ gradually refines the automatic discrete tag set, giving intuitive fine-grained clusters of English words. 

\subsection{Tradeoff Curves} 
\label{sub:rate_distortion_curve}
\begin{figure}[t]
\centering
\subfloat[Discrete Version]{ \includegraphics[page=1,width=0.45\textwidth]{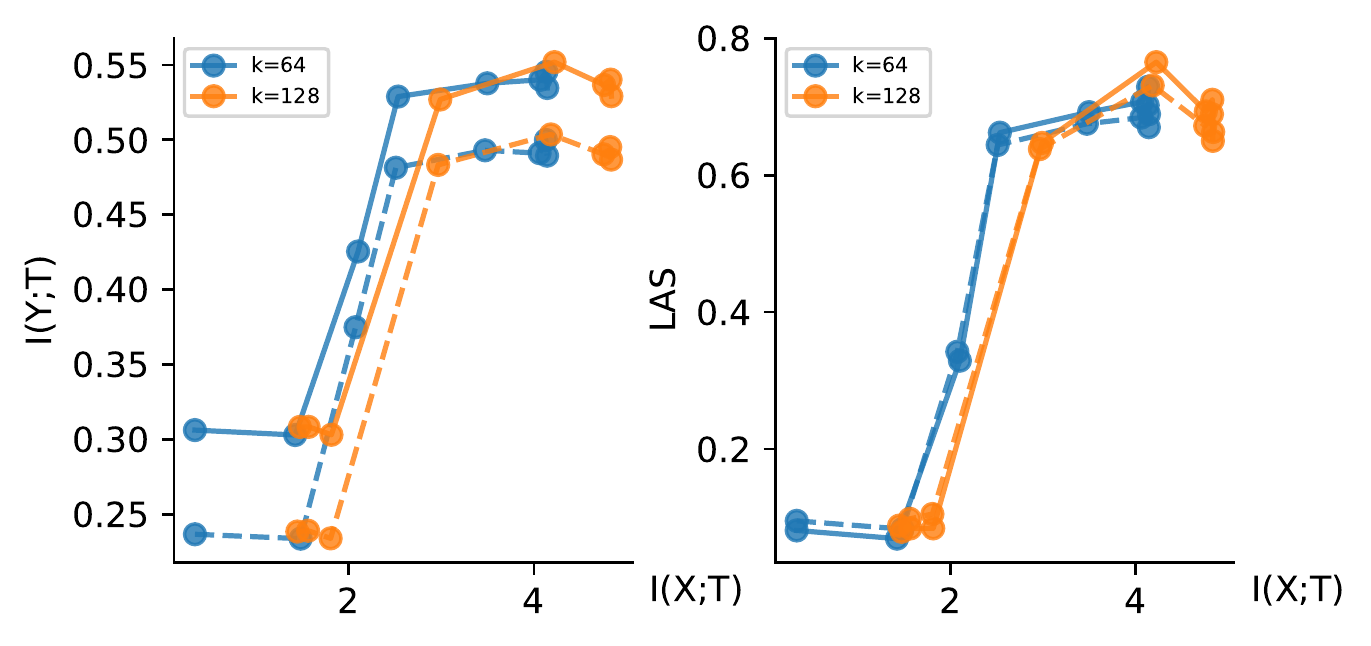}}
\\
\subfloat[Continuous Version]{ \includegraphics[page=1,width=0.45\textwidth]{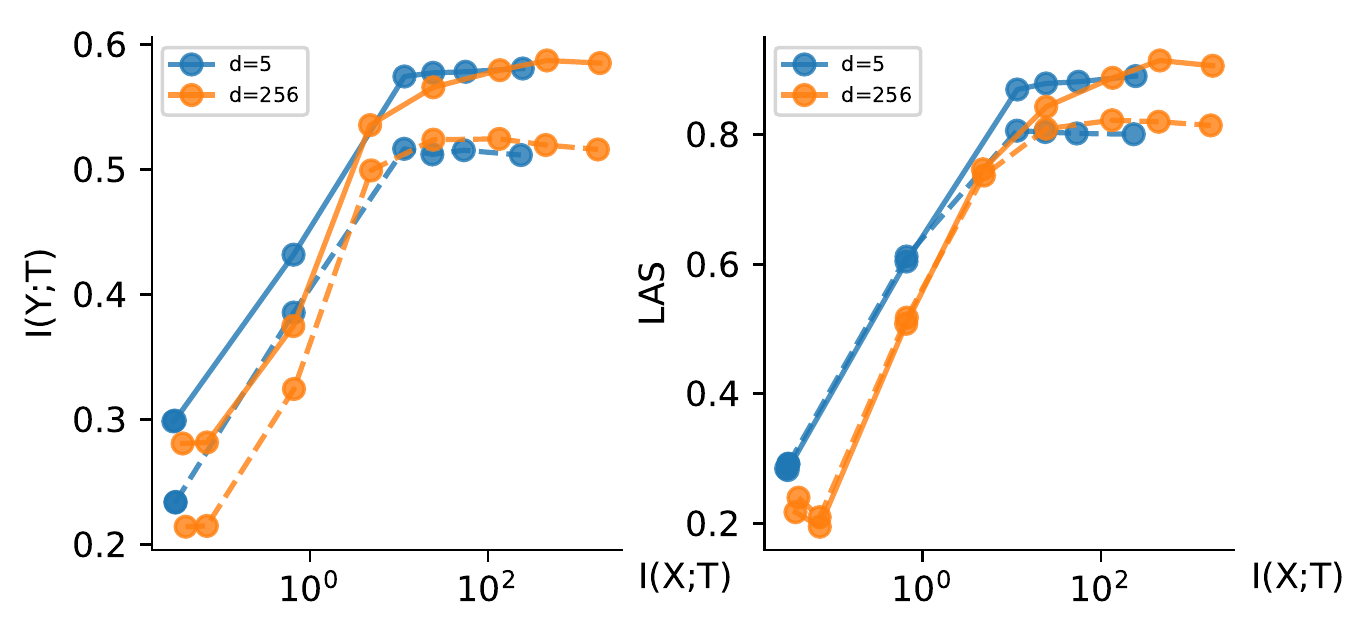}}
\caption{\label{fig:rate_distortion} Compression-prediction tradeoff curves of VIB in our dependency parsing setting. The upper figures use discrete tags, while the lower figures use continuous tags. The dashed lines are for test data, and the solid lines for training data.  The ``dim'' in the legends means the dimensionality of the continuous tag vector or the cardinality of the discrete tag set. 
On the left, we plot predictiveness $\MI(Y;T)$ versus $\MI(X;T)$ as we lower $\beta$ multiplicatively from $10^1$ to $10^{-6}$ on a log-scale. On the right, we alter the $y$-axis to show the labeled attachment score (LAS) of 1-best dependency parsing.  All mutual information and entropy values in this paper are reported in nats per token.  Furthermore, the mutual information values that we report are actually our variational upper bounds, as described in \cref{sec:formal_model}. The reason that $\MI(X;T)$ is so large for continuous tags is that it is \emph{differential} mutual information (see \cref{fn:diffent}). 
Additional tradeoff curves w.r.t. $\MI(T_i;X \mid \xType)$ are in \cref{supp:tradeoff}.}
\end{figure}

As we lower $\beta$ to retain more information about $X$, both $\MI(X;T)$ and $\MI(Y;T)$ rise, as shown in \cref{fig:rate_distortion}.  There are diminishing returns: after some point, the additional information retained in $T$ does not contribute much to predicting $Y$.  
 Also noteworthy is that at each level of $\MI(X,T)$, very low-dimensional tags ($d=5$) perform on par with high-dimensional ones ($d=256$). (Note that the high-dimensional stochastic tags will be noisier to keep the same $\MI(X,T)$.)  The low-dimensional tags allow far faster CPU parsing.
 This indicates that VIB can achieve strong practical task-specific compression.
 
\begin{figure*}[ht!]
\centering
\setlength\tabcolsep{2pt}
\centering
\resizebox{0.8\textwidth}{!}{
  \hspace{-30pt}
  \subfloat[ELMo, $\MI(X;T)=H(X) \approx 400.6$]{ \includegraphics[page=3,width=0.33\textwidth]{fig/cluster/visual_5k_test_toomuch.pdf} }
  \hspace{-20pt}
\qquad
\subfloat[$\MI(X;T)\approx24.3$]{\includegraphics[page=2,width=0.33\textwidth]{fig/cluster/visual_5k_test_toomuch.pdf}}
\qquad
  \hspace{-20pt}
\subfloat[$\MI(X;T)\approx 0.069$]{ \includegraphics[page=1,width=0.33\textwidth]{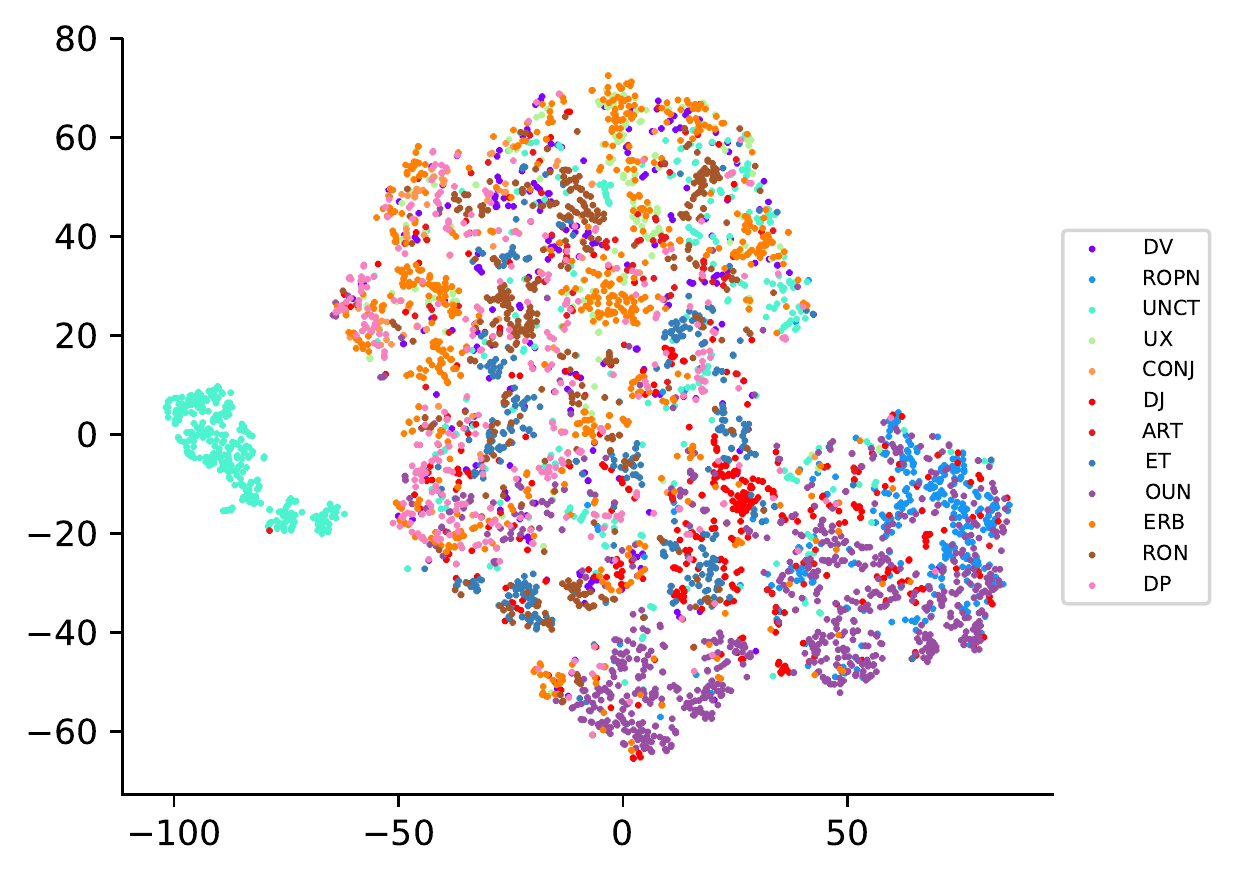} }
  \hspace{-30pt}
  \qquad
  \hspace{20pt}
 \includegraphics[width=0.10\textwidth]{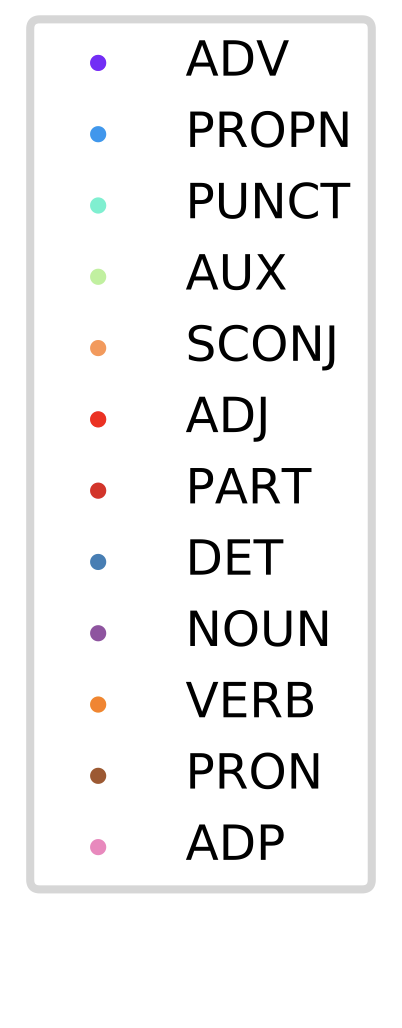}
  \hspace{-30pt}}

\caption{\label{fig:visual_pos} t-SNE visualization of  VIB model ($d=256$) on the projected space of the continuous tags. Each marker in the figure represents a word token, colored by its gold POS tag. This series of figures (from left to right) shows a progression from no compression to moderate compression and to too-much compression.}
\end{figure*}

\subsection{Learned Tags vs.\@ Gold POS Tags} 
We investigate how our automatic tag $T_i$ correlates with the gold POS tag $A_i$ provided by UD.
\label{sub:pos_tags_analysis}
\paragraph{Continuous Version}
We use t-SNE \cite{tsne} to visualize our compressed continuous tags on held-out test data, coloring each token in \cref{fig:visual_pos} \lisa{make the legend bigger} \lisa{DONE} according to its gold POS tag.  (Similar plots for the discrete tags are in \cref{figsupp:app-visual_pos} in the appendix.)

In \cref{fig:visual_pos}, the first figure shows the original uncompressed level-1 ELMo embeddings of the tokens in test data.  In the two-dimensional visualization, the POS tags are vaguely clustered but the boundaries merge together and some tags are diffuse. The second figure is when $\beta=10^{-3}$ (moderate compression): our compressed embeddings show clear clusters that correspond well to gold POS tags.  Note that the gold POS tags were not used in training either ELMo or our method.  The third figure is when $\beta=1$ (too much compression), when POS information is largely lost. An interesting observation is that the purple \texttt{NOUN} and blue \texttt{PROPN} distributions overlap in the middle distribution, meaning that it was unnecessary to distinguish common nouns from proper nouns for purposes of our parsing task.\footnote{Both can serve as arguments of verbs and prepositions.  Both can be modified by determiners and adjectives, giving rise to proper NPs like ``The Daily Tribune.''}

\paragraph{Discrete Version}
We also quantify how well our specialized discrete tags capture the traditional POS categories, by investigating $\MI(A; T)$.  This can be written as $\Entr(A) - \Entr(A \mid T)$.  Similarly to \cref{sec:decoder}, our probability distribution has the form $\ptheta(x,a,t) = p(x,a) \cdot \ptheta(t\mid x)$, leading us to write $\Entr(A \mid T) \leq \E[x,a]{\E[t \sim \ptheta(t \mid x)]{-\log q(a \mid t)}}$ where $q(a \mid t) = \prod_i q(a_i \mid t_i)$ is a variational distribution that we train to minimize this upper bound.  This is equivalent to training $q(a \mid t)$ by maximum conditional likelihood.
In effect, we are doing transfer learning, fixing our trained IB encoder ($\ptheta$) and now using it to predict $A$ instead of $Y$, but otherwise following \cref{sec:decoder}.
We similarly upper-bound $\Entr(A)$ by assuming a model $q'(a) = \prod_i q'(a_i)$ and estimating $q'$ as the empirical distribution over training tags.  Having trained $q$ and $q'$ on training data, we estimate $\Entr(A\mid T)$ and $\Entr(A)$ using the same upper-bound formulas on our test data.

\begin{figure*}
\begin{minipage}{\textwidth}
\begin{minipage}[t]{0.2\textwidth}
  \centering
  \vspace{0pt}
  \includegraphics[page=1,width=\textwidth]{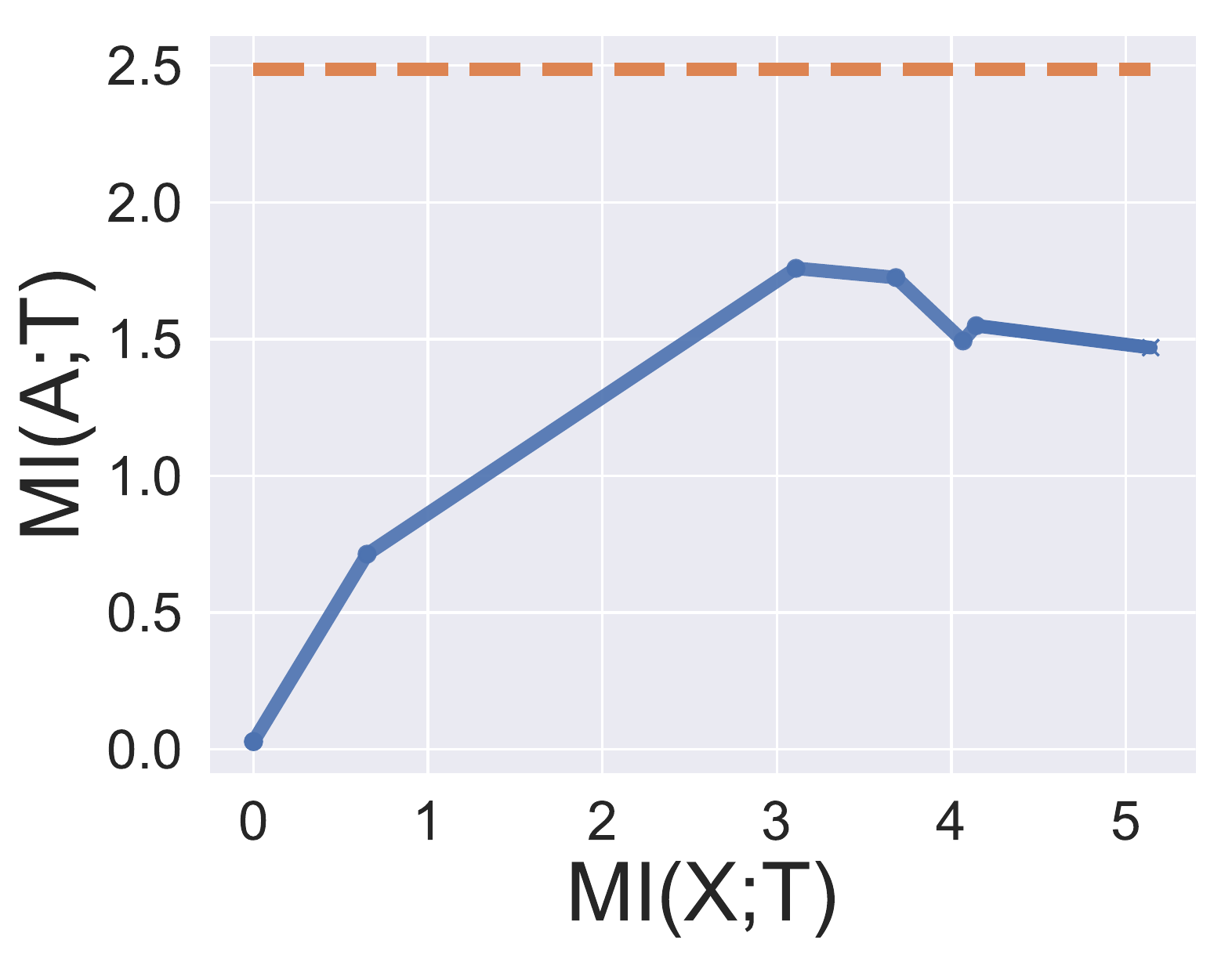}
\end{minipage}
  \hfill
\begin{minipage}[t]{0.8\textwidth}
\centering
\vspace{3pt}
\resizebox{\textwidth}{!}{
\begin{tabular}{lllllllllll}
\toprule
embeddings  &Arabic &English &Spanish &French &Hindi &Italian &Portuguese &Russian &Chinese \\
$\Entr(A)$  &2.016   &2.486   &2.345   &2.206  &2.247  &2.291  &2.306  &2.131  &2.195 \\ \hline
ELMo0           &67.2\%  &74.2\%   &75.7\%    &79.6\%   &70.1\%   &77.9\%   &76.5\%   &73.2\%   &57.3\%\\
ELMo1           &67.2\%  &76.1\%   &71.7\%   &78.0\%   &70.5\%   &78.1\%   &72.3\%   &73.8\%   &59.8\%\\
ELMo2           &63.8\%  &71.0\%   &79.7\%   &78.7\%   &67.2\%   &74.5\%   &75.3\%   &72.2\%   &59.4\%\\
\bottomrule 

\end{tabular}}
\end{minipage}
\end{minipage}
\caption{\label{fig:few_shot} Graph at left: $\MI(A;T)$ vs.\@ $\MI(X;T)$ in English (in units of nats per token). \jason{there's a commented-out reference to another graph here, because that graph was buggy---fix?} 
Table at right: how well the discrete specialized tags predict gold POS tags for 9 languages. The $\Entr(A)$ row is the entropy (in nats per token) of the gold POS tags in the test data corpus, which is an upper bound for $\MI(A;T)$. The remaining rows report the percentage $\MI(A;T) / \Entr(A)$.}
\vspace{2mm}
\end{figure*}

We  experiment on all 9 languages, taking $T_i$ at the moderate compression level $\beta=0.001$, $k=64$. 
As  \cref{fig:few_shot} shows, averaging over the $9$ languages, the reconstruction retains $71\%$ of POS information (and as high as 80\% on Spanish and French). We can conclude that the information encoded in the specialized tags correlates with the gold POS tags, but does not perfectly predict the POS. 

The graph in \cref{fig:few_shot} shows a ``U-shaped'' curve, with the best overall error rate at $\beta=0.01$. That is, moderate compression of ELMo embeddings helps for predicting POS tags. Too much compression squeezes out POS-related information, while too little compression allows the tagger to overfit the training data, harming generalization to test data. We will see the same pattern for parsing in \cref{sec:extrinsic}.

\paragraph{Syntactic Features}
As a quick check, we determine that our tags also make syntactic distinctions beyond those that are recognized by the UD POS tag set, such as tense, number, and transitivity.  See \cref{supp:subcat} for graphs.  For example, even with moderate compression, we achieve $0.87$ classification accuracy in distinguishing between transitive and intransitive English verbs, given only tag $t_i$.

\paragraph{Stem} When we compress ELMo embeddings to $k$ discrete tags, the semantic information must be squeezed out because $k$ is small. But what about the continuous case? In order to verify that semantic information is excluded, we train a classifier that predicts the stem of word token $i$ from its mean tag vector $\E{T_i}$. We expect ``player'' and ``buyer'' to have similar compressed vectors, because they share syntactic roles, but we should \emph{fail} to predict that they have different stems ``play'' and ``buy.'' 
The classifier is a feedforward neural network with $tanh$ activation function, and the last layer is a softmax over the stem vocabulary. In the English treebank, we take the word lemma in UD treebank and use the NLTK library \cite{nltk} to stem each lemma token. Our result (\cref{supp:stem} in the appendix) suggests that more compression destroys stem information, as hoped.  With light compression, the error rate on stem prediction can be below $15\%$.  With moderate compression $\beta=0.01$, the error rate is $89\%$ for ELMo layer 2 and $66\%$ for ELMo layer 1. Other languages show the same pattern, as shown in \cref{supp:stem} in the appendix. Thus, moderate and heavy compression indeed squeeze out semantic information. 

\subsection{Annealing of Discrete Tags}\label{sec:da}
Deterministic annealing \cite{rose-1998,Friedman:2001} is a method that gradually decreases $\beta$ during training of IB.
Each token $i$ has a stochastic distribution over the possible tags $\{1,\ldots,k\}$.  This can be regarded as a soft clustering where each token is fractionally associated with each of the $k$ clusters.  With high $\beta$, the optimal solution turns out to assign to all tokens an identical distribution over clusters, for a mutual information of 0.  Since all clusters then have the same membership, this is equivalent to having a single cluster.  As we gradually reduce $\beta$, the cluster eventually splits.  Further reduction of $\beta$ leads to recursive splitting, yielding a hierarchical clustering of tokens (\cref{sec:detanneal}).

We apply deterministic annealing to the English dataset, and the resulting hierarchical structure reflects properties of English syntax. At the top of the hierarchy, the model places nouns, adjectives, adverbs, and verbs in different clusters.  At lower levels, the anaphors (``yourself,'' ``herself'' \dots), possessive pronouns (``his,'' ``my,'' ``their'' \dots), accusative-case pronouns (``them,''  ``me,'' ``him,'' ``myself'' \dots), and nominative-case pronouns (``I,'' ``they,'' ``we'' \dots) each form a cluster, as do the \emph{wh}-words (``why,'' ``how,'' ``which,'' ``who,'' ``what,'' \dots).  \jason{add picture} \lisa{I think the description we have here is already very clear. Adding a figure would be re-stating the same thing.}

\begin{table*}[t]
\centering\small
\begin{tabular}{llllllllllll}
\toprule
& Models    & Arabic  & Hindi    & English & French  & Spanish   &Portuguese  & Russian &  Chinese & Italian\\


\toprule 
&\texttt{Iden} & 0.751    & \textbf{0.870}   & 0.824   & 0.784   & 0.808     & 0.813   & 0.783   & 0.709 & \textbf{0.863}\\
&\texttt{PCA} & 0.743 & \textbf{0.866} &0.823& 0.749 & 0.802 & 0.808 & 0.777 & 0.697 & 0.857 \\
&\texttt{MLP} & 0.759    & \textbf{0.871}   & 0.839   & 0.816   & \textbf{0.835}     & 0.821   & 0.800   & 0.734 & \textbf{0.867}\\
&\texttt{VIBc}   & \textbf{0.779}   & \textbf{0.866}   & \textbf{0.851}   &  \textbf{0.828}  &  \textbf{0.837}    &\textbf{0.836}   & \textbf{0.814}   &  \textbf{0.754}&\textbf{0.867}\\
\midrule
&\texttt{\grey{POS}} & \grey{ 0.652}  & \grey{0.713} & \grey{0.712} & \grey{0.718} & \grey{\textbf{0.739}} & \grey{\textbf{0.743}} & \grey{\textbf{0.662}} & \grey{0.510 } & \grey{0.779} \\
 &\texttt{\grey{VIBd}} &  \grey{\textbf{0.672}} &  \grey{\textbf{0.736}}  &  \grey{\textbf{0.742}} &  \grey{\textbf{0.723}} & \grey{ \textbf{0.725}} &  \grey{0.710} &  \grey{\textbf{0.651}} &  \grey{\textbf{0.591}} &  \grey{\textbf{0.781}} \\
\bottomrule

\end{tabular}
\caption{\label{tb:multilingual} Parsing accuracy of 9 languages (LAS).  Black rows use continuous tags; gray rows use discrete tags (which does worse).  In each column, the best score for each color is boldfaced, along with all results of that color that are not significantly worse (paired permutation test, $p < 0.05$).
These results use only ELMo layer 1; results from all layers are shown in \cref{tb:app-multilingual} in the appendix, for both LAS and UAS metrics.}
\end{table*}

\section{Engineering Evaluation} 
\label{sec:extrinsic}

As we noted in \cref{sec:intro}, learning how to 
compress ELMo's tags for a given task is a fast alternative to fine-tuning all the ELMo parameters. \lisa{todo: compare to the baseline proposed by reviewer3, which is to add a cleanup layer after the lstm layer. }  \lisa{DONE}
We find that indeed, training a compression method to keep only the relevant information does improve our generalization performance on the parsing task.

We compare 6 different token representations according to the test accuracy of a dependency parser trained to use them.  The same training data is used to jointly train the parser and the token encoder that produces the parser's input representations. 
\\
\textbf{Continuous tags:} \\
\textbf{\texttt{Iden}} is an baseline model that leaves the ELMo embeddings uncompressed, so $d=1024$. \\
\textbf{\texttt{PCA}} is a baseline that simply uses Principal Components Analysis to reduce the dimensionality to $d=256$.  Again, this is not task-specific. \\
\textbf{\texttt{MLP}} is another deterministic baseline that uses a multi-layer perceptron (as in \citet{DBLP:journals/corr/DozatM16})
to reduce the dimensionality to $d=256$ in a task-specific and nonlinear way.  This is identical to our continuous VIB method except that the variance of the output Gaussians is fixed to 0, so that the $d$ dimensions are fully informative.\\
\textbf{\texttt{VIBc}} uses our stochastic encoder, still with $d=256$.  The average amount of stochastic noise is controlled by $\beta$, which is tuned per-language on dev data. \\
\textbf{Discrete tags:} \\
\textbf{\texttt{POS}} is a baseline that uses the $k \leq 17$ gold POS tags from the UD dataset. \\
\textbf{\texttt{VIBd}} is our stochastic method with $k=64$ tags. To compare fairly with \textbf{\texttt{POS}}, we pick a $\beta$ value for each language such that $\Entr(T_i \mid X_i) \approx \Entr(A_i \mid X_i)$.

\paragraph{Runtime.} Our VIB approach is quite fast. With minibatching on a single GPU, it is able to train on 10,000 sentences in 100 seconds, per epoch.

\paragraph{Analysis.} \Cref{tb:multilingual} shows the test accuracies of these parsers,
using the standard training/development/test split for each UD language.   

In the continuous case, the VIB representation outperforms all three baselines in 8 of 9 languages, and is not significantly worse in the 9th language (Hindi). In short, our VIB joint training generalizes better to test data.  This is because the training objective \eqref{eq:Obj3} includes terms that focus on the parsing task and also regularize the representations. 

In the discrete case, the VIB representation outperforms gold POS tags (at the same level of granularity) in 6 of 9 languages, and of the other 3, it is not significantly worse in 2.  This suggests that our learned discrete tag set could be an improved alternative to gold POS tags  \cite[cf.][]{klein-manning-2003-unlex} when a discrete tag set is needed for speed.




\section{Related Work}
\label{sec:related}
Much recent NLP literature examines syntactic information encoded by deep models \cite{Linzen-lstm-syntax} and more specifically, by powerful unsupervised word embeddings. \newcite{hewitt2019structural} learn a linear projection from the embedding space to predict the distance between two words in a parse tree. \newcite{BiLM-dissect} and \newcite{BERT-syntax} assess the ability of BERT and ELMo directly on syntactic NLP tasks. \newcite{tenney2019probe} extract information from the contextual embeddings by self-attention pooling within a span of word embeddings. 

The IB framework was first used in NLP to cluster distributionally similar words \cite{Pereira:1993}.  In cognitive science, it has been used to argue that color-naming systems across languages are nearly optimal \cite{color}. In machine learning, IB provides an information-theoretic perspective to explain the performance of deep neural networks \cite{Tishby2015DeepLA}.

The VIB method makes use of variational upper and lower bounds on mutual information.  An alternative lower bound was proposed by \newcite{mi_bound:Poole}, who found it to work better empirically.



\section{Conclusion and Future Work} 
\label{sec:future_work}
In this paper, we have proposed two ways to syntactically compress ELMo word token embeddings, using variational information bottleneck.  We automatically induce stochastic discrete tags that correlate with gold POS tags but are as good or better for parsing.  
We also induce stochastic continuous token embeddings (each is a Gaussian distribution over $\mathbb{R}^d$) that forget non-syntactic information captured by ELMo. These stochastic vectors yield improved parsing results, in a way that simpler dimensionality reduction methods do not.  They also transfer to the problem of predicting gold POS tags, which were not used in training. 

One could apply the same training method to compress the ELMo or BERT token sequence $x$ for other tasks.  All that is required is a model-specific decoder $\qphi(y\mid t)$.  For example, in the case of sentiment analysis, the approach should preserve only sentiment information, discarding most of the syntax.  One possibility that does not require supervised data is to create artificial tasks, such as reproducing the input sentence or predicting missing parts of the input (such as affixes and function words).  In this case, the latent representations would be essentially generative, as in the variational autoencoder \cite{kingma2013auto}.

 
\section*{Acknowledgments} 
This work was supported by the National Science Foundation under Grant No.\@  1718846 and by a Provost's Undergraduate Research Award to the first author.  The Maryland Advanced Research Computing Center provided computing facilities.  We thank the anonymous reviewers and Hongyuan Mei for helpful comments.\par

\bibliographystyle{acl_natbib}
\bibliography{acl2019}
\clearpage\appendix\appendixpage

\section{Details of Deterministic Annealing}\label{sec:detanneal}
In practice, deterministic annealing (\cref{sec:da}) is implemented in a way that dynamically increases the number of clusters $k$ \cite{Friedman:2001}, leading to a hierarchical clustering.
First, we initialize with one cluster, and all the word tokens are mapped to that cluster with probability 1. Second, for each cluster $i$, duplicate the cluster $C_i$ to form $C_{ia}, C_{ib}$, and divide the probabilities associated with $C_i$ approximately evenly (with perturbation) between the two clusters, i.e., set $p(c_{ia} | x)  = \frac{1}{2} p(c_i | x) + \epsilon_x$ and $p(c_{ib} | x)  = \frac{1}{2} p(c_i | x) - \epsilon_x$. Third, update $\beta \leftarrow \beta/\alpha$, and run optimization until convergence. Fourth, for each former cluster $i$, if $C_{ia}$ and $C_{ib}$ have not differentiated from each other, re-merge them by setting $p(c_i | x) = p(c_{ia} | x) + p(c_{ib} | x)$.  (Optimization will have pulled them together again for higher $\beta$ values and pushed them apart for lower $\beta$ values.)  Our heuristic is to re-merge them if for all word tokens $x$, $|p(c_{ia} | x) - p(c_{ib} | x)| \leq 0.01$.  
Finally, loop back to the second step, unless the $\beta$ value has fallen below a given threshold $\beta_{\min}$ or we have reached a desired maximum number of clusters. 

\section{Additional Tradeoff Curves}\label{supp:tradeoff}
\begin{figure*}[]
\subfloat[Discrete Version -- ELMo layer-1]{ \includegraphics[page=1,width=0.5\textwidth]{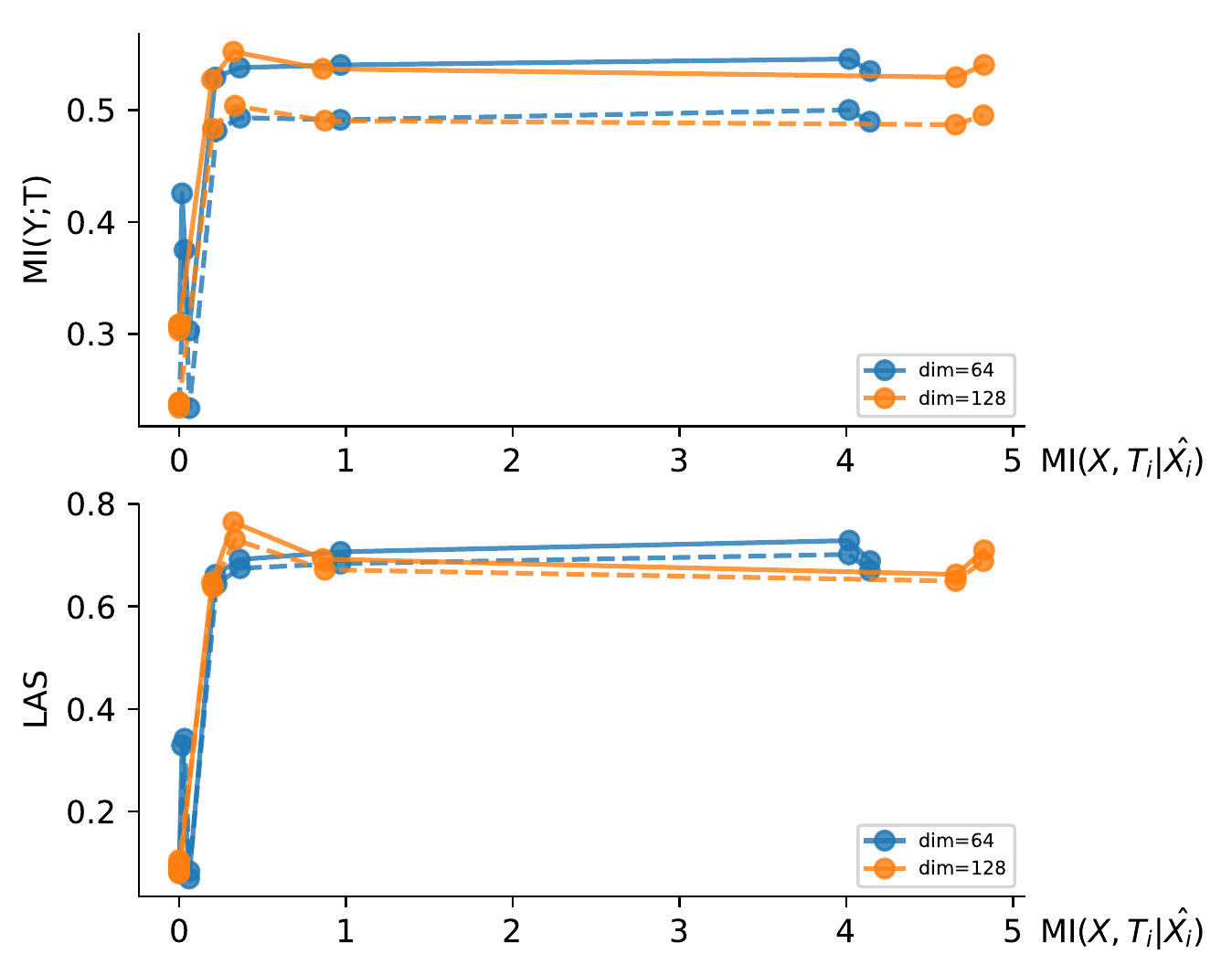} }
\subfloat[Continuous Version -- ELMo layer-1]{ \includegraphics[page=1,width=0.5\textwidth]{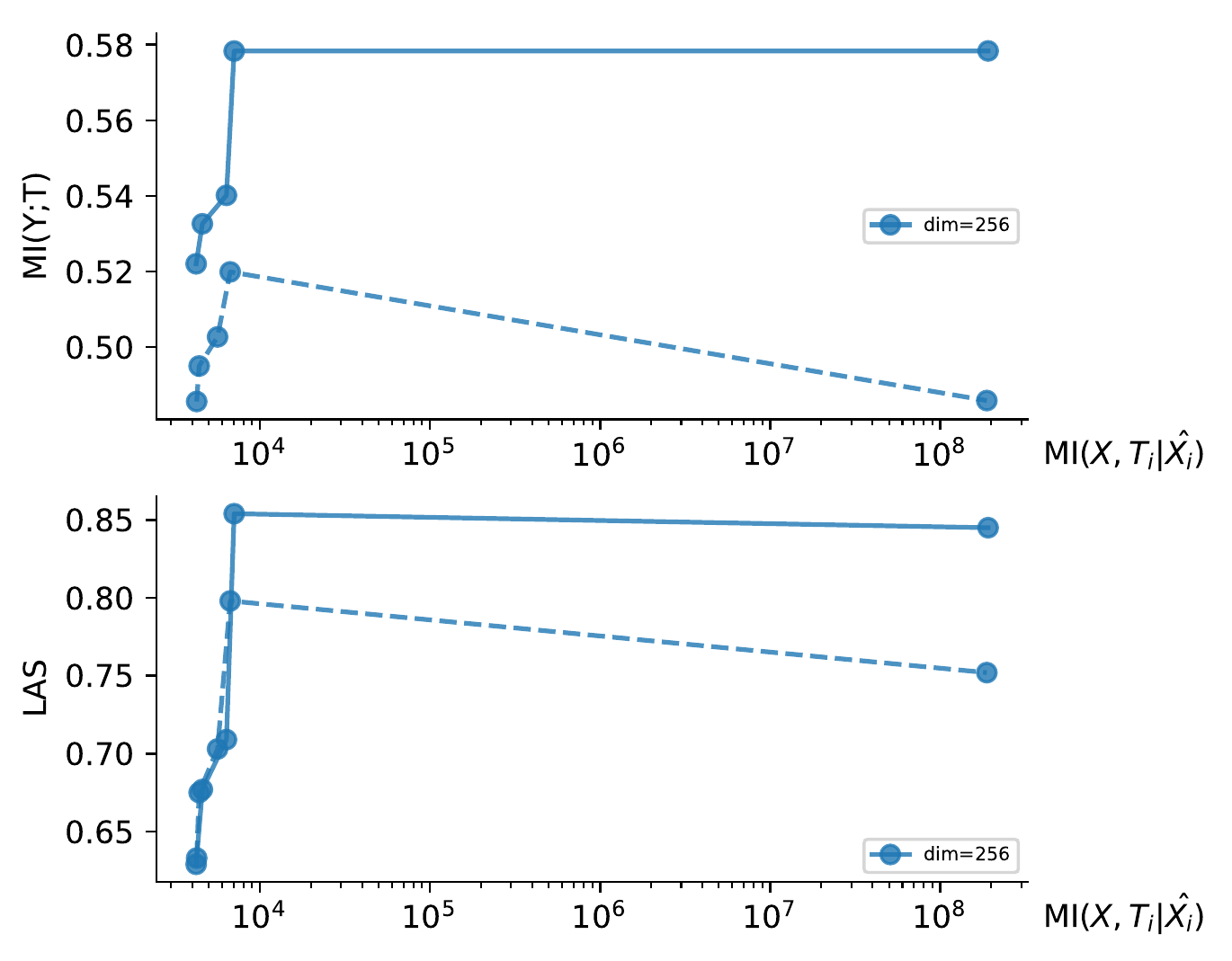} }
\caption{\label{figsupp:app-type_tradeoff}. Tradeoff curves for $\MI(T_i;X \mid \xType)$ vs.\@ $\MI(Y;T)$ and $\MI(T_i;X \mid \xType)$ vs.\@ LAS, complementary to \cref{fig:rate_distortion}. }
\end{figure*}

\cref{figsupp:app-type_tradeoff} supplements the tradeoff curves in \cref{fig:rate_distortion} by plotting the relationship between $\MI(T_i;X \mid \xType)$ vs.\@ $\MI(Y;T)$, and $\MI(T_i;X \mid \xType)$ vs.\@  LAS.   Moving leftward on the graphs, each $T_i$ contains less \emph{contextual} information about word $i$ (because $\gamma$ in \cref{eq:Obj3} is larger) as well as less information overall about word $i$ (because we always set $\gamma=\beta$, so $\beta$ is larger as well).  The graphs show that the tag sequence $T$ then becomes less informative about the parse $Y$.

\section{Additional t-SNE plots}\label{supp:tsne}

\begin{figure*}[]
\centering
  \hspace{-30pt}
  \subfloat[ELMo, Continuous, Types]{ \includegraphics[page=1,width=0.33\textwidth]{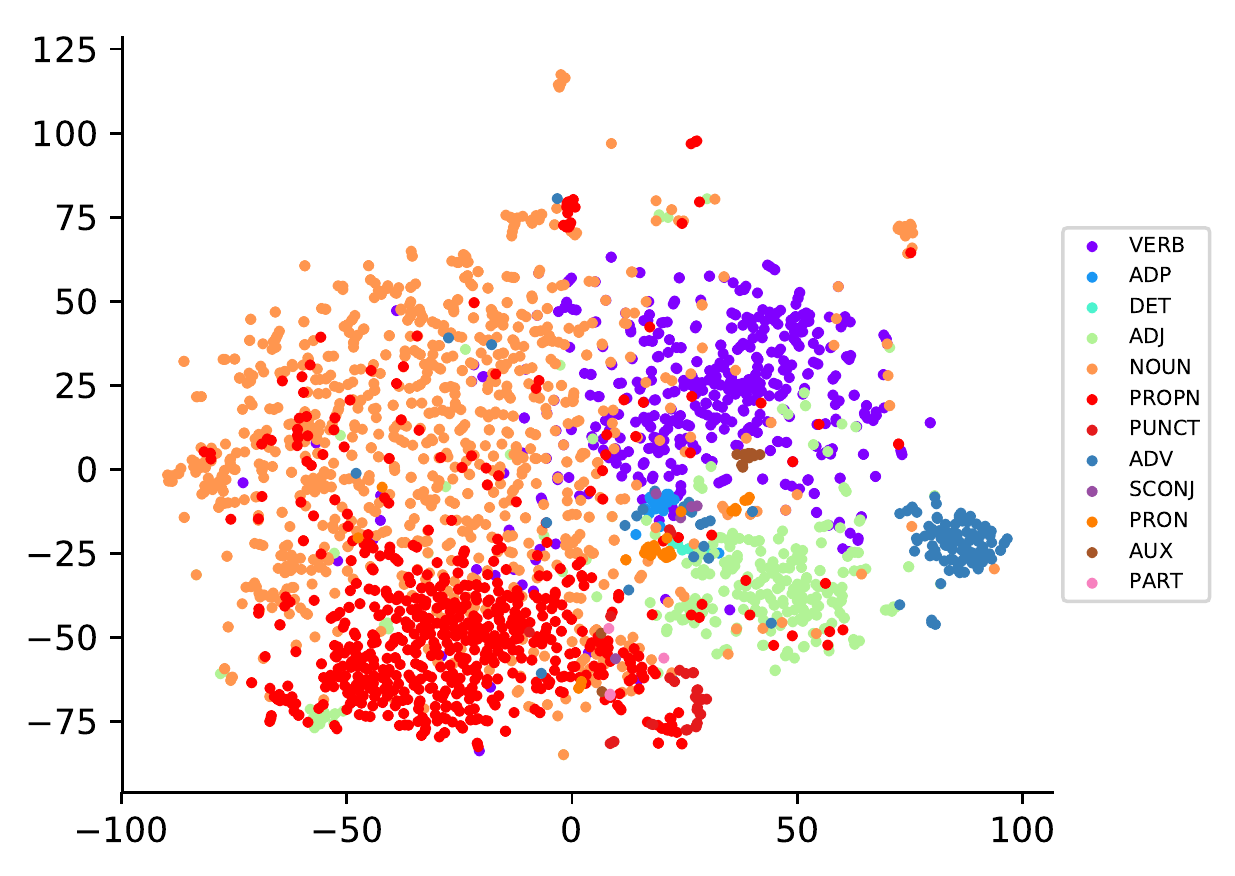} }
  \hspace{-20pt}
\qquad
\subfloat[$\MI(X;T)\approx123.4$]{\includegraphics[page=1,width=0.33\textwidth]{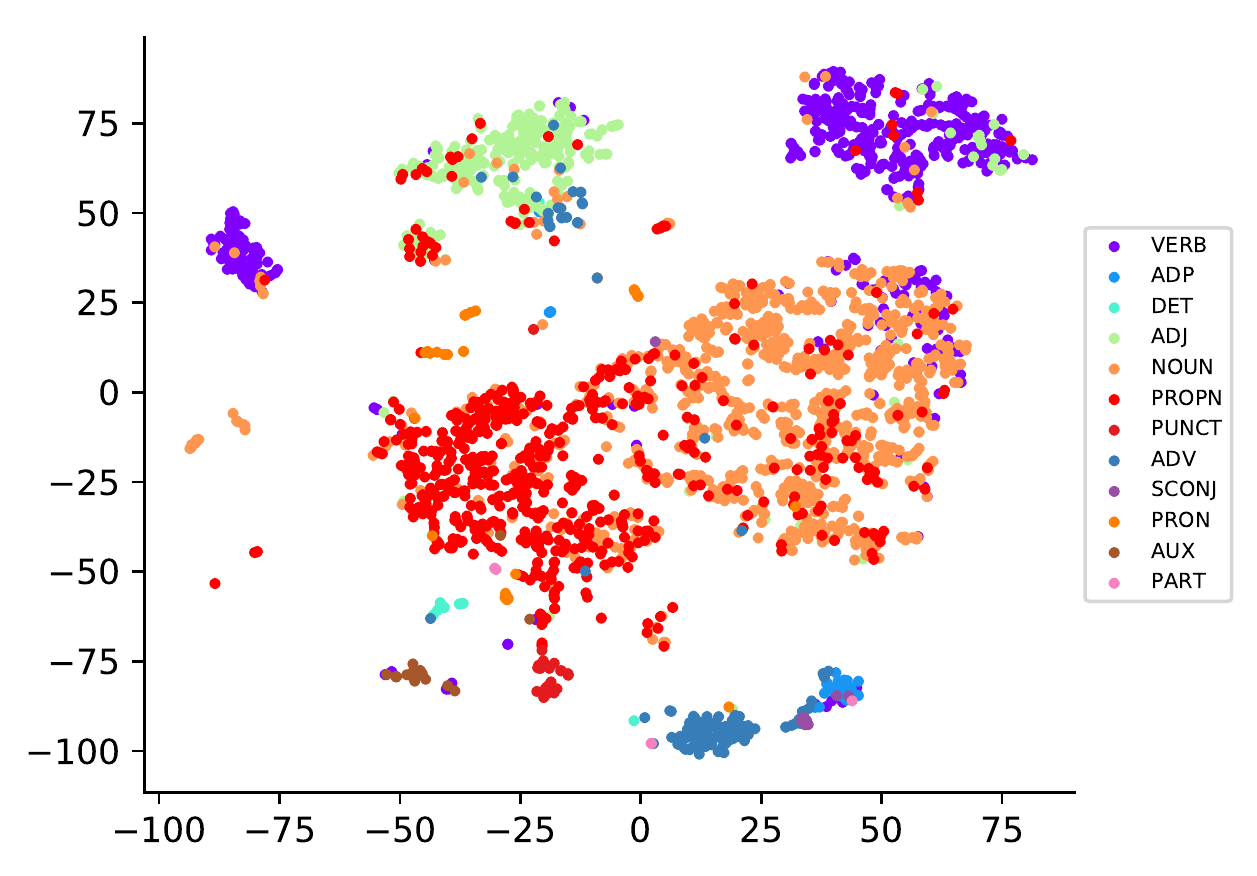}}
\qquad
  \hspace{-20pt}
\subfloat[$\MI(X;T)\approx 0.333$]{ \includegraphics[page=1,width=0.33\textwidth]{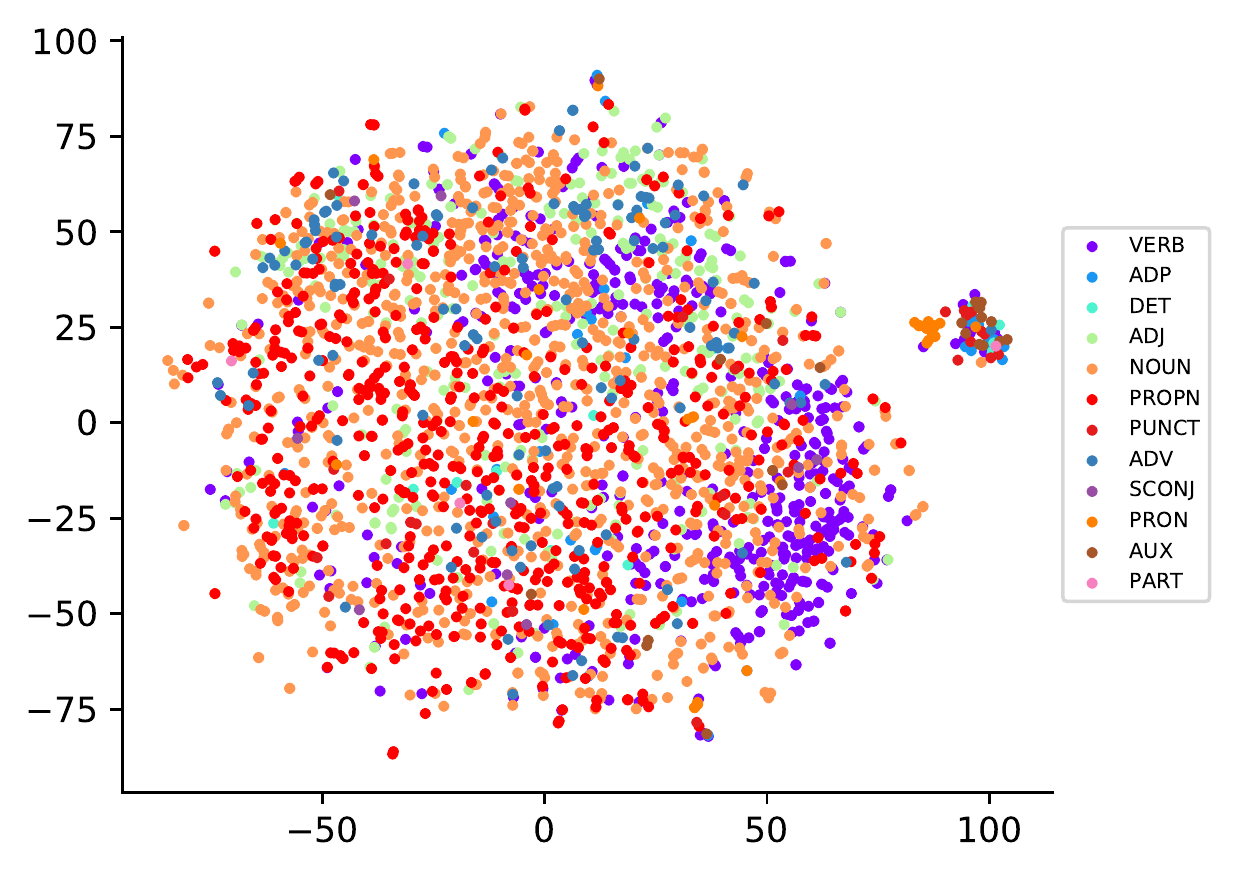} }\\
  \hspace{-30pt}
   \subfloat[ELMo, Discrete, Types]{ \includegraphics[page=1,width=0.33\textwidth]{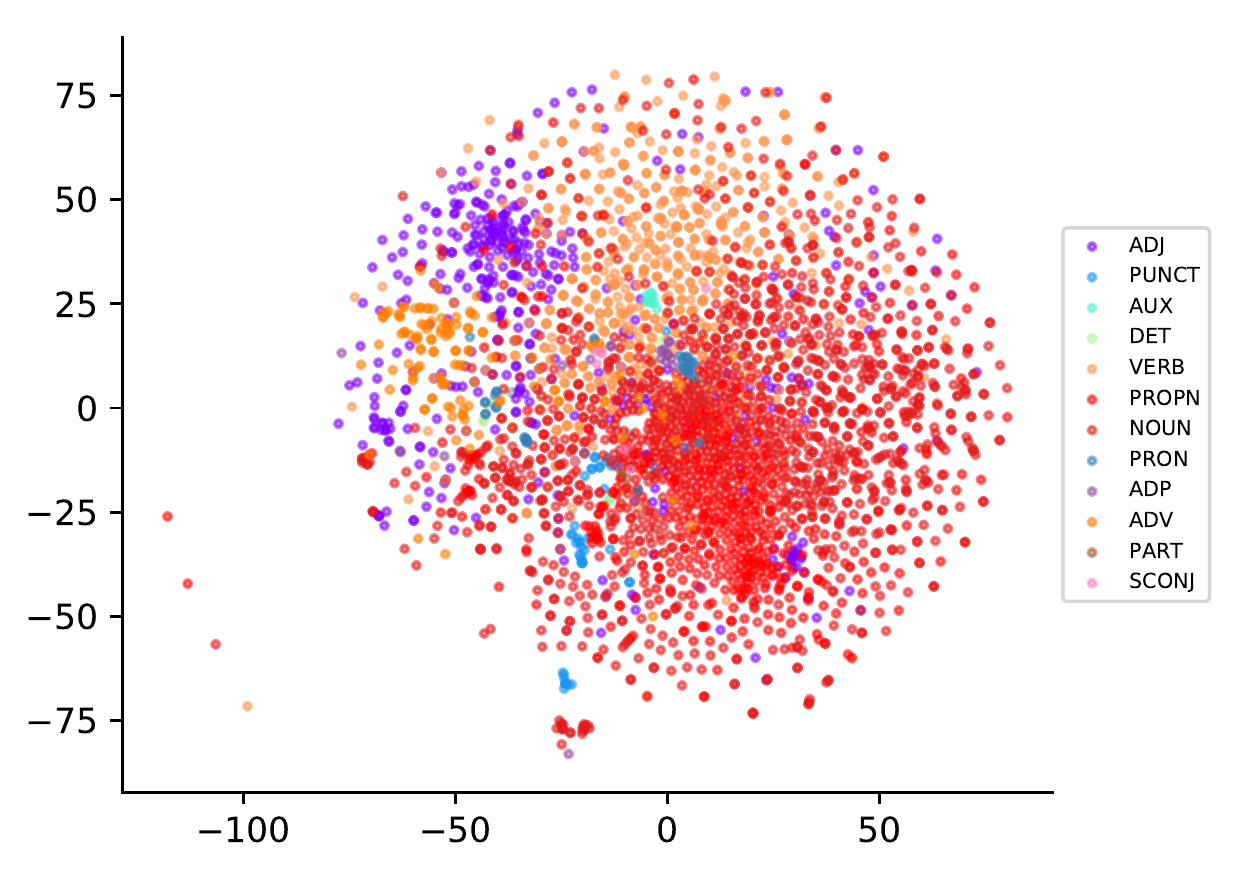} }
  \hspace{-30pt}
\qquad
  \subfloat[$\MI(X;T) \approx 3.958$]{ \includegraphics[page=1,width=0.33\textwidth]{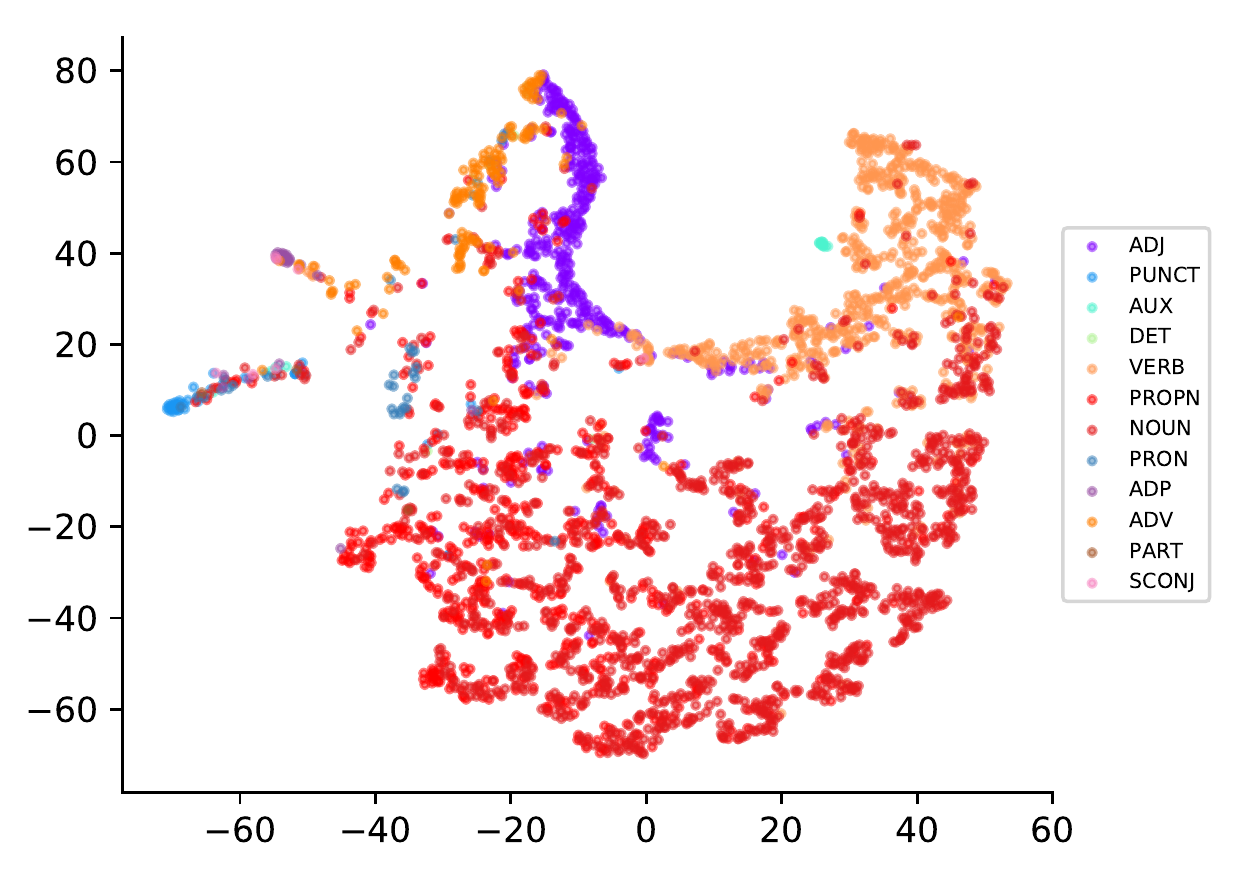} }
\qquad
  \hspace{-20pt}
    \subfloat[$\MI(X;T) \approx 1.516$]{ \includegraphics[page=1,width=0.33\textwidth]{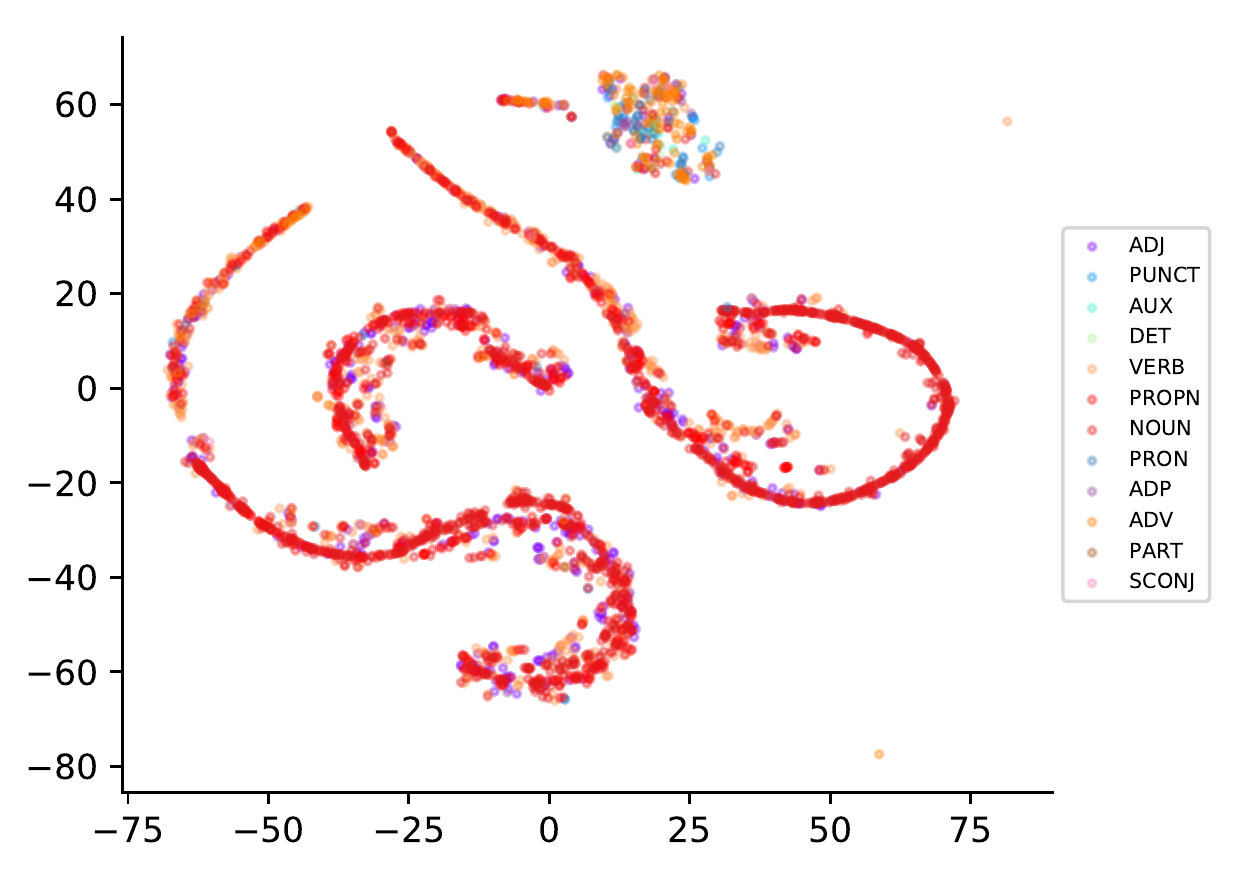} }\\


 \hspace{-30pt}
  \subfloat[ELMo, Discrete, Tokens]{ \includegraphics[page=1,width=0.33\textwidth]{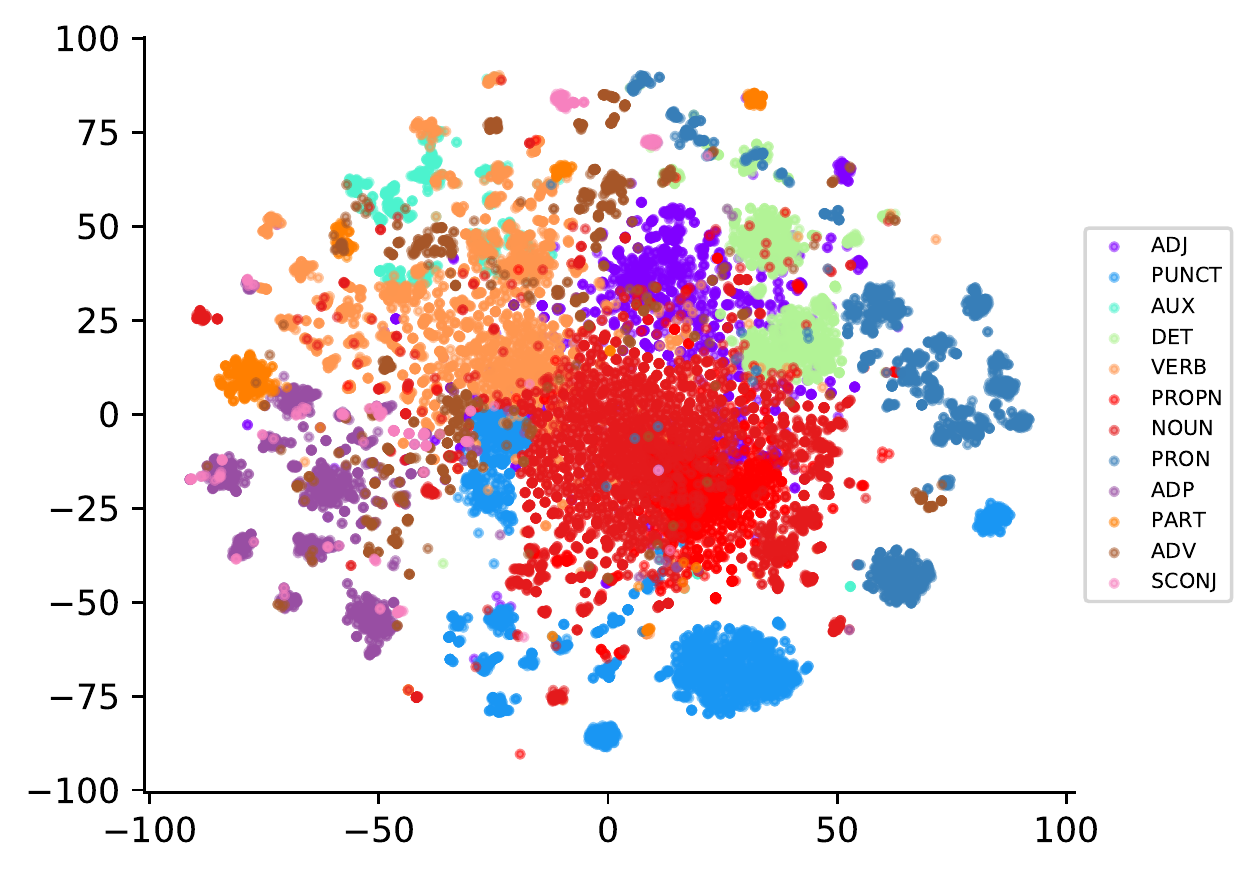} }
  \qquad
  \hspace{-30pt}
  \subfloat[$\MI(X;T) \approx 4.755$]{ \includegraphics[page=1,width=0.33\textwidth]{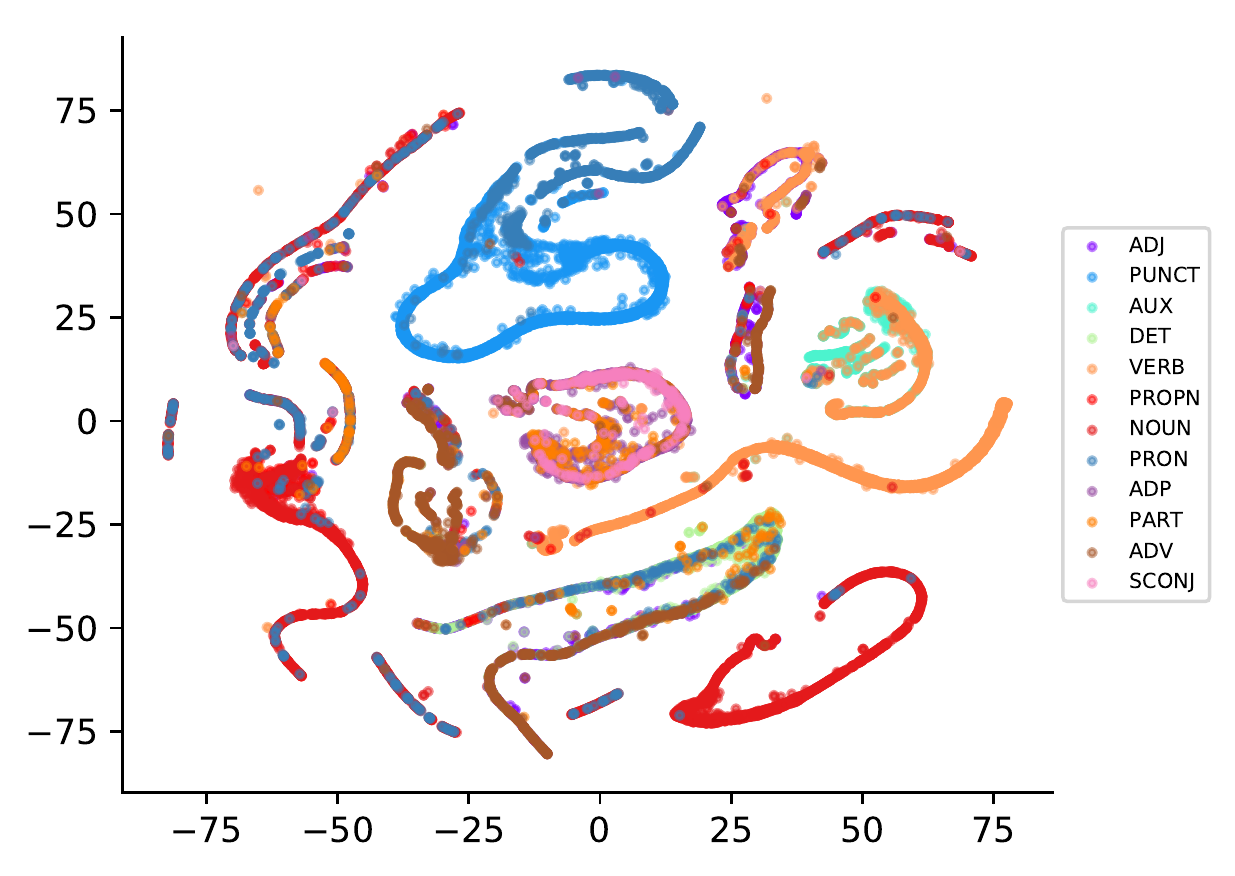} }
  \qquad
  \hspace{-20pt}
    \subfloat[$\MI(X;T) \approx 1.475$]{ \includegraphics[page=1,width=0.33\textwidth]{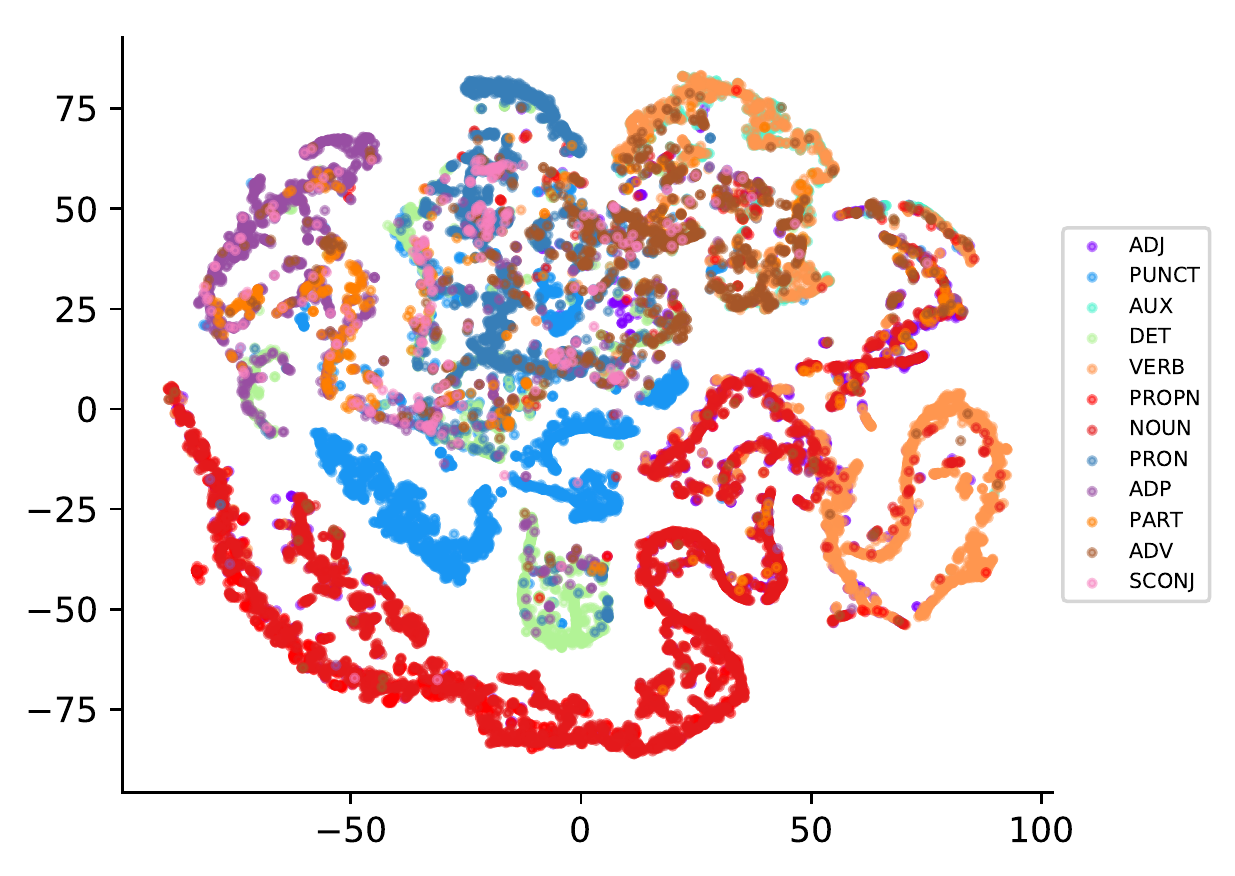} }

\caption{\label{figsupp:app-visual_pos} t-SNE visualization of our continuous tags ($d=256$) and our distributions over discrete tags ($k=128$), supplementing \cref{fig:visual_pos}. Each marker in the figure represents a word token, colored by its gold POS tag. For each row, the series of figures (from left to right) shows a transition from no compression to moderate compression and to too-much compression. The first row (a-c) shows the continuous type embeddings; the second row (d-f) shows the discrete type embeddings; the third row (g-i) shows the discrete token embeddings. }

\end{figure*}

Recall that \cref{fig:visual_pos} (in 
\cref{sub:pos_tags_analysis}) was a row of t-SNE visualizations of the continuous \emph{token} embeddings $\ptheta(t_i \mid x_i)$ under no compression, moderate compression, and too much compression.  \Cref{figsupp:app-visual_pos} gives another row visualizing the continuous \emph{type} embeddings $\sxi(t_i \mid \xtype)$ in the same way.  In both cases, the ``moderate compression'' condition shows $\beta=0.01$.

\Cref{figsupp:app-visual_pos} also shows rows for the \emph{discrete} type and token embeddings.  In both cases, the ``moderate compression'' condition shows $\beta=0.001$.

In the continuous case, each point given to t-SNE is the mean of a Gaussian-distributed stochastic embedding, so it is in $\mathbb{R}^d$.  In the discrete case, each point given to t-SNE is a vector of $k$ tag probabilities, so it is in $\mathbb{R}^k$ and more specifically in the $(k-1)$-dimensional simplex.  The t-SNE visualizer plots these points in 2 dimensions.

The message of all these graphs is that the tokens or types with the same gold part of speech (shown as having the same color) are most nicely grouped together in the moderate compression condition.


\section{Syntactic Feature Classification}
\label{supp:subcat}
\begin{figure*}[ht!]
\centering
\includegraphics[page=1,width=0.5\textwidth]{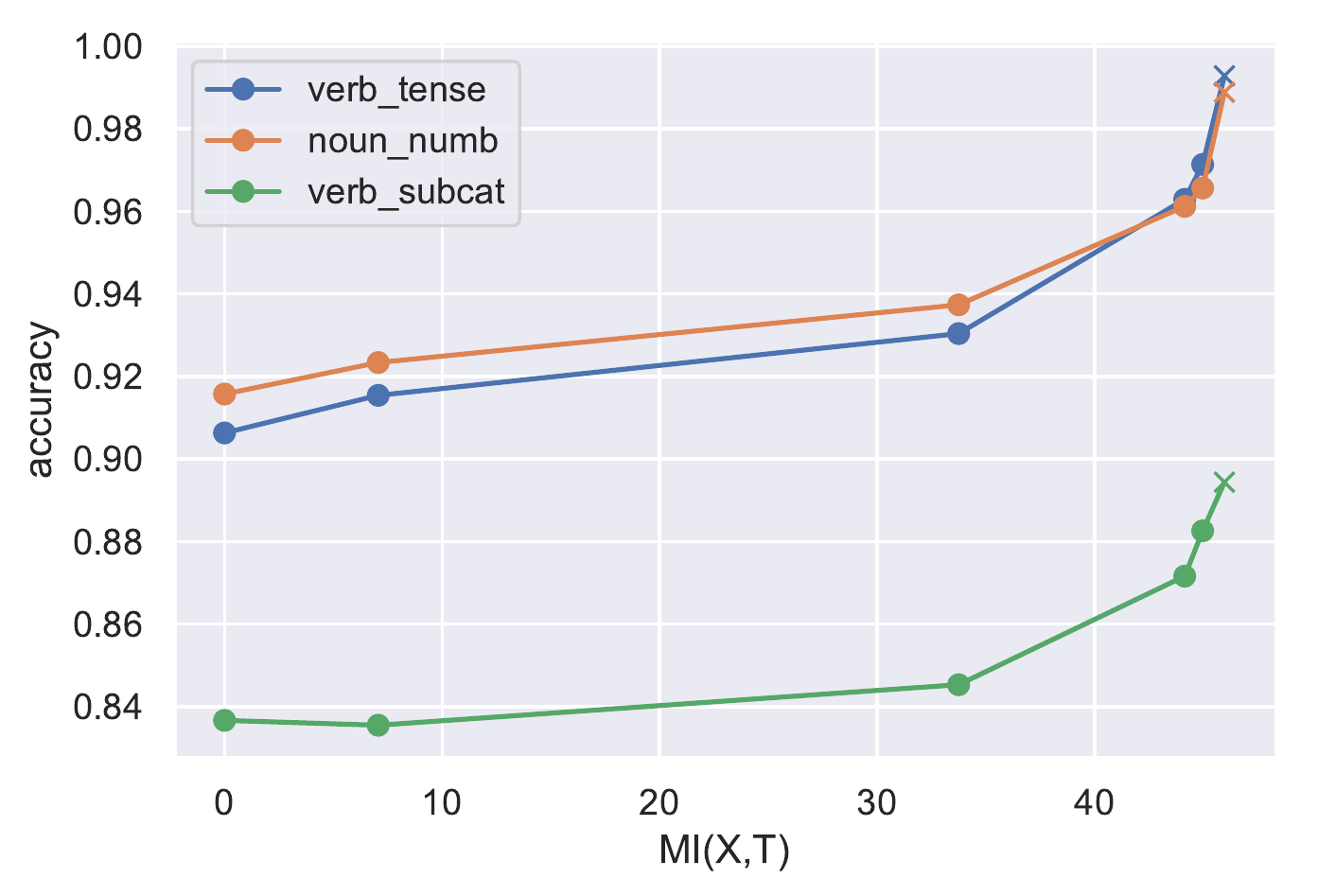}
\caption{\label{figsupp:subcat} The accuracy of predicting the subcategorization frame of verbs (transitive/intransitive), number of nouns (plural/singular), and tense of verbs (past/present/future), as we change the level of compression of ELMo layer-1 (see the \textbf{Subcategorization frame} paragraph in \cref{sub:pos_tags_analysis}).  As we move from right to left and squeeze irrelevant information out of the tags, they retain these three syntactic distinctions quite well.}
\end{figure*}

\Cref{figsupp:subcat} shows results for the \textbf{Syntactic Features} paragraph in \cref{sub:pos_tags_analysis}, by showing the prediction accuracy of subcategorization frame, tense, and number from $t_i$ as a function of  the level of compression.  We used an SVM classifier with a radial basis function kernel.

All results are on the English UD data with the usual training/test split.  To train and test the classifiers, we used the gold UD annotations to identify the nouns and verbs and their correct syntactic features.  

\section{Plot \& Table for Stem Prediction}
\label{supp:stem}
\begin{figure*}[h]
\centering
\includegraphics[page=1,width=0.5\textwidth]{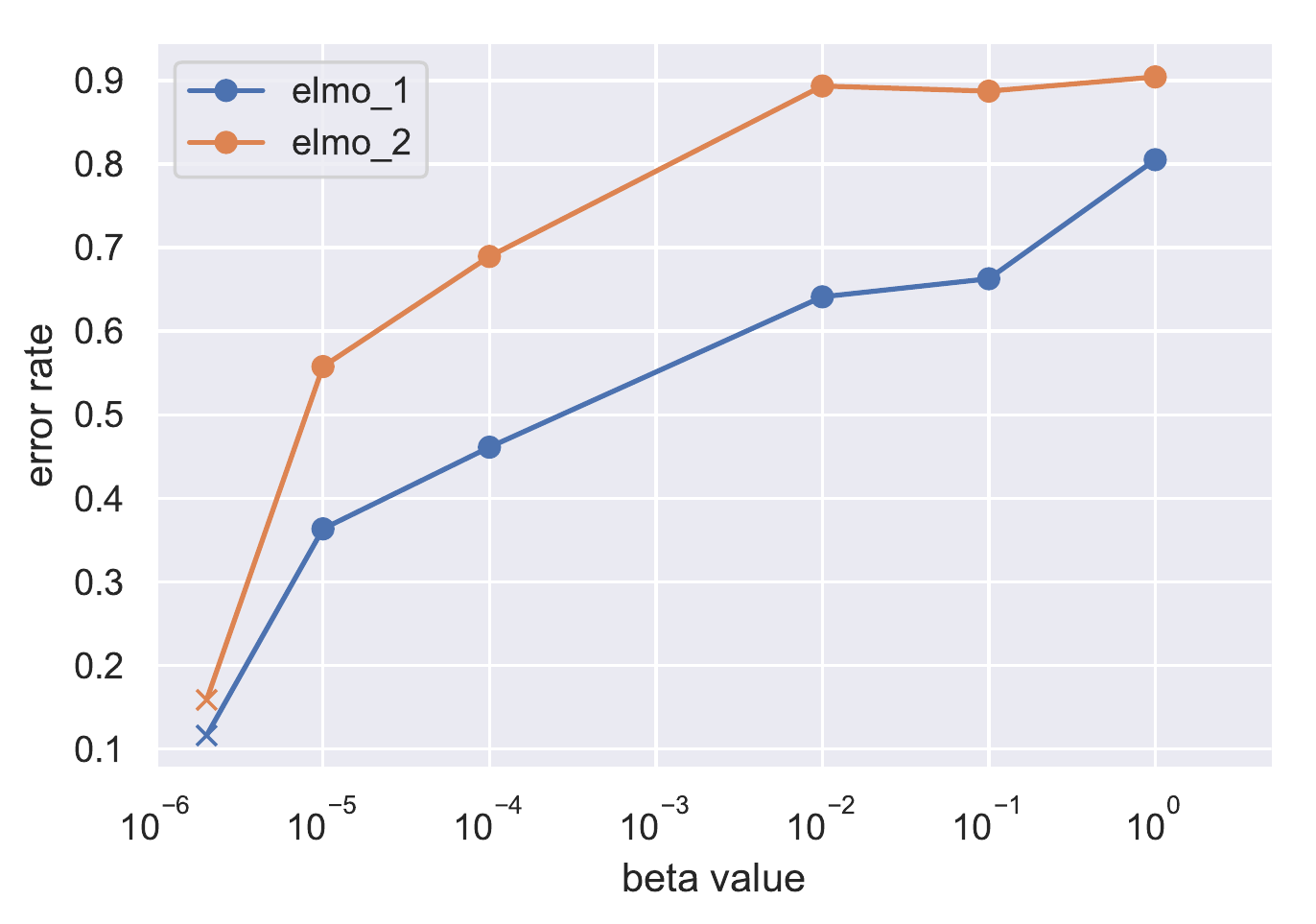}
\caption{\label{figsupp:recon_sem} The error rate of reconstructing the stem of a word from the specialized continuous tags. The legend indicates whether we are compressing the ELMo layer-1 or ELMo layer-2 (see the \textbf{Stem} paragraph of \cref{sub:pos_tags_analysis}).
}\Jason{would have been better to use MI on the x axis}
\end{figure*}

\Crefrange{figsupp:recon_sem}{figsupp:semforeign} supplement the \textbf{Stem} paragraph in \cref{sub:pos_tags_analysis}. \Cref{figsupp:recon_sem} plots the error rate of reconstructing English stems as a function of the level of compression. \Cref{figsupp:semforeign} shows the reconstruction error rate for the other 8 languages. 
\section{Additional Table of Parsing Performance}
\Cref{tb:app-multilingual} is an extended version of \cref{tb:multilingual} in \cref{sec:extrinsic}.  It includes parsing performance (measured by LAS and UAS) using ELMo layer 0, 1, and 2. 

\begin{figure*}
\centering
\vspace{3pt}
\resizebox{\textwidth}{!}{
\begin{tabular}{llllllllll}
\toprule
Compression  & layer      &Arabic  &Spanish     &French   &Hindi    &Italian   &Portuguese &Russian &Chinese \\
Slight       & 1          &26.7\%   &24.0\%     &25.5\%   &20.5\%   &29.9\%       &32.5\%   &38.8\%   &26.7\%\\
Moderate     & 1          &89.5\%   &79.8\%     &66.7\%   &94.6\%   &94.7\%       &93.7\%   &94.0\%   &89.6\%\\

Slight       & 2          &34.9\% &34.9\% &34.9\% &34.9\% &34.9\% &34.9\% &34.9\% &34.9\% \\
Moderate     & 2          &94.3\% &94.3\% &94.3\% &94.3\% &94.3\% &94.3\% &94.3\% &94.3\% \\
\bottomrule 

\end{tabular}}
\caption{\label{figsupp:semforeign} Error rate in reconstructing the stem of a word from the compressed version of the ELMo layer-1 and layer-2 embedding). Slight compression refers to  $\beta=0.0001$, and moderate compression refers to $\beta=0.01$.}
\Jason{should perhaps add level 1 for completeness, as in Fig 9 and section 6.2} \lisa{DONE}
\end{figure*}

\begin{table*}[ht!]
\centering
\setlength\tabcolsep{2pt}
\resizebox{0.8\textwidth}{!}{
\begin{tabular}{llllllllllll}
\toprule
\multicolumn{11}{c}{UAS}  \\
& Models &Layer    & Arabic  & Hindi        & English & French  & Spanish   &Portuguese  & Russian &  Chinese & Italian\\
\midrule
&Iden &0            &0.817 & 0.914          & 0.793 & 0.836     & 0.851     & 0.844      & 0.859  & 0.775 & 0.904\\
&Iden &1            &0.821 & \textbf{0.915} & 0.868 & 0.833     & 0.852     & 0.842      & 0.860  &  0.771  & 0.903\\
&Iden &2            &0.820 & 0.914          & 0.843 & 0.833     & 0.856     & 0.841      & 0.859  & 0.773 & 0.901\\

& PCA & 0           &0.814  & 0.912         &0.787  & 0.814     & 0.847     & 0.857      & 0.831 &  0.773 & 0.897 \\
& PCA & 1           &0.815  & 0.912         &0.865  & 0.807     & 0.846     & 0.855      & 0.828 &  0.759 & 0.899 \\
& PCA & 2           &0.814  & 0.915         &0.832  & 0.808     & 0.846     & 0.858      & 0.829 &  0.766 & 0.902 \\

& MLP           & 0 &0.830  &0.918 &0.742 &0.856  &0.829  &0.869  &0.852  &0.797  &0.910 \\

& MLP           & 1 &0.831  &0.923 &0.823  &0.870  &0.832  &0.867  &0.852  &0.800  &0.908 \\

& MLP           & 2 &0.833  &0.918 &0.787 &0.859  &0.813  &0.871  &0.849  &0.790  &\textbf{0.914}    \\
&VIBc           &0  &0.852 & \textbf{0.915} & 0.866 & \textbf{0.879} & \textbf{0.881} & 0.871 & 0.862 &  0.800 & 0.831\\
&VIBc &1  & \textbf{0.860}   & \textbf{0.913}   & \textbf{0.871}   & \textbf{0.877}  &  \textbf{0.880}    & \textbf{0.877}   & \textbf{0.865}   & \textbf{0.814} & \textbf{0.913}\\
&VIBc &2  & 0.851   & 0.894     & \textbf{0.880}   &  0.876  &  0.879    & \textbf{0.877}   & 0.843  & 0.768 & 0.878\\
\midrule
&\grey{POS} &- & \grey{0.722}  & \grey{0.819} & \grey{0.762}  & \grey{0.800} & \grey{0.802} & \grey{\textbf{0.808}} & \grey{0.739} &  \grey{0.570} & \grey{\textbf{0.843}} \\
&\grey{VIBd} &\grey{0} & \grey{\textbf{0.783}} &\grey{0.823}  &\grey{0.784} &\grey{\textbf{0.821}}  &\grey{\textbf{0.821}}  &\grey{0.793}  &\grey{\textbf{0.777}}  &\grey{0.671}  &\grey{\textbf{0.855}} \\ 
&\grey{VIBd} &\grey{1} & \grey{\textbf{0.784}} & \grey{\textbf{0.862}}  & \grey{\textbf{0.825}} & \grey{\textbf{0.822}} & \grey{\textbf{0.822}} & \grey{\textbf{0.805}} & \grey{\textbf{0.776}} &  \grey{\textbf{0.691}} & \grey{\textbf{0.857}} \\  
&\grey{VIBd} &\grey{2} & \grey{0.754} &\grey{\textbf{0.861}}  &\grey{0.816} &\grey{\textbf{0.822}}  &\grey{0.812}  &\grey{0.790}  &\grey{0.768}  &\grey{0.672}  &\grey{0.849}\\

\bottomrule
\toprule 
\multicolumn{11}{c}{LAS}  \\
  & Models      &layer & Arabic             & Hindi           & English & French  & Spanish   & Portuguese  & Russian   &  Chinese  & Italian\\
&Iden             &0  &0.747                & \textbf{0.867}  & 0.745   & 0.789   & 0.806     & 0.812       & 0.788     & 0.713     &\textbf{0.864}\\
&Iden             &1  &0.751                & \textbf{0.870}  & 0.824   & 0.784   & 0.808     & 0.813       & 0.783     & 0.709     & \textbf{0.863}\\
&Iden             &2  &0.743                & 0.867           & 0.798   & 0.782   & 0.811     & 0.813       &0.787      & 0.713     &0.861\\
& PCA             & 0 &0.746                & 0.864           &0.742    & 0.758   & 0.804     & 0.811       & 0.781     &  0.706    & 0.856 \\
& PCA             & 1 & 0.743               & \textbf{0.866}  &0.823    & 0.749   & 0.802     & 0.808       & 0.777     &  0.697    & 0.857 \\
& PCA             & 2 &0.744                & \textbf{0.870}  &0.787    & 0.750   & 0.801     & 0.811       & 0.780     &  0.700    & \textbf{0.865} \\

& MLP             & 0 &0.754 & \textbf{0.869}  &0.801  &0.814  &0.772  &0.817  &0.798  &0.739  &\textbf{0.871} \\
& MLP             & 1 & 0.759               & \textbf{0.871}  &0.839    & 0.816   &\textbf{0.835}& 0.821    & 0.800   & 0.734 & \textbf{0.867}\\
& MLP             & 2 &0.760 &\textbf{0.871}  &0.834 &0.814  &0.755  &0.822  &0.797  &0.726  &0.869\\


&VIBc             &0  & \textbf{0.778}     & 0.865            & 0.822  &  0.822  &  \textbf{0.839}    & 0.827   & 0.807   &  0.739& 0.862\\
&VIBc             &1  & \textbf{0.779}      & \textbf{0.866}  & \textbf{0.851}   &  \textbf{0.828}  &  \textbf{0.837}    &\textbf{0.836}   & \textbf{0.814}   &  \textbf{0.754}&\textbf{0.867}\\
&VIBc  &2  & 0.777   & 0.838   & 0.840   &  \textbf{0.826}    &  0.840    & 0.829   & 0.786    &0.710&0.818\\
\midrule
& \grey{POS} &- & \grey{0.652}  & \grey{0.713} & \grey{0.712} & \grey{0.718} & \grey{\textbf{0.739}} & \grey{\textbf{0.743}} & \grey{\textbf{0.662}} & \grey{0.510 } & \grey{0.779} \\
 & \grey{VIBd} &\grey{0}& \grey{\textbf{0.671}} &\grey{0.702} &\grey{0.721}  &\grey{\textbf{0.723}}  &\grey{\textbf{0.724}}  &\grey{0.710}  &\grey{0.648}  &\grey{0.544}  &\grey{\textbf{0.780} } \\
 & \grey{VIBd} &\grey{1}&  \grey{\textbf{0.672}} &  \grey{\textbf{0.736}}   &  \grey{\textbf{0.742}} &  \grey{\textbf{0.723}} & \grey{ \textbf{0.725}} &  \grey{0.710} &  \grey{\textbf{0.651}} &  \grey{\textbf{0.591}} &  \grey{\textbf{0.781}} \\
 & \grey{VIBd} &\grey{2}&\grey{0.643} &\grey{\textbf{0.735}}  &\grey{\textbf{0.741}} &\grey{0.721}  &\grey{0.719}  &\grey{0.698}  &\grey{0.646}  &\grey{0.566}  &\grey{0.763} \\
\bottomrule

\end{tabular}}
\caption{\label{tb:app-multilingual} Parsing accuracy of 9 languages (LAS and UAS);  \cref{tb:multilingual} is a subset of this table.  Black rows use continuous tags; gray rows use discrete tags (which does worse). The ``layer'' column indicates the ELMo layer we use. In each column, the best score for each color is boldfaced, along with all results of that color that are not significantly worse (paired permutation test, $p < 0.05$). }
\end{table*}




\end{document}